\documentclass[11pt,reqno]{amsart}
\usepackage{float} 
\usepackage{amsthm}
\usepackage{cases}
\usepackage{amscd}
\usepackage{amsfonts}

\usepackage{amssymb}
\usepackage{amsmath}
\usepackage[dvipdfmx]{graphicx}
\usepackage{booktabs}
\usepackage{tabularx}
\usepackage{caption} 
\usepackage{cleveref}
\usepackage{url}

\usepackage{subfigure}
\usepackage{algorithm}
\usepackage{algpseudocode} 
\renewcommand{\algorithmicrequire}

\usepackage{cleveref}
\usepackage{color}
\usepackage{datetime}

\allowdisplaybreaks

\makeatletter
\newcommand\figcaption{\def\@captype{figure}\caption}
\newcommand\tabcaption{\def\@captype{table}\caption}
\makeatother

\usepackage{geometry}
\usepackage{algorithm}
\usepackage{multirow}
\usepackage{mathrsfs}

\newtheorem{remark}{Remark}

\newcommand{\beq}{\begin{equation}}
\newcommand{\eeq}{\end{equation}}
\newcommand{\bea}{\begin{eqnarray}}
\newcommand{\eea}{\end{eqnarray}}
\newcommand{\beas}{\begin{eqnarray*}}
\newcommand{\eeas}{\end{eqnarray*}}

\begin{document}
\title{Weights initialization of neural networks for function approximation}

\author[X. Hu, Y. Huang, N. Yi, and P. Yin]{Xinwen Hu$^\dagger$, Yunqing Huang$^{\S}$, Nianyu Yi$^{\dagger, \ddagger}$, Peimeng Yin$^*$}
\address{$^\dagger$ School of Mathematics and Computational Science, Xiangtan University, Xiangtan 411105, P.R.China }
\email{xinwen618@gmail.com}
\address{$\S$ National Center for Applied Mathematics in Hunan, Key Laboratory of Intelligent Computing \& Information Processing of Ministry of Education, Xiangtan University, Xiangtan 411105, Hunan, P.R.China} 
\email{huangyq@xtu.edu.cn}
\address{$^\ddagger$ Hunan Key Laboratory for Computation and Simulation in Science and Engineering, Xiangtan University, Xiangtan 411105, P.R.China}
\email{yinianyu@xtu.edu.cn}
\address{$^*$ Department of Mathematical Sciences, University of Texas at El Paso,  El Paso, Texas 79968, USA}
\email{pyin@utep.edu}



\begin{abstract}
Neural network-based function approximation plays a pivotal role in the advancement of scientific computing and machine learning. Yet, training such models faces several challenges: (i) each target function often requires training a new model from scratch; (ii) performance is highly sensitive to architectural and hyperparameter choices; and (iii) models frequently generalize poorly beyond the training domain.
To overcome these challenges, we propose a reusable initialization framework based on basis function pretraining. In this approach, basis neural networks are first trained to approximate families of polynomials on a reference domain. Their learned parameters are then used to initialize networks for more complex target functions. To enhance adaptability across arbitrary domains, we further introduce a domain mapping mechanism that transforms inputs into the reference domain, thereby preserving structural correspondence with the pretrained models.
Extensive numerical experiments in one- and two-dimensional settings demonstrate substantial improvements in training efficiency, generalization, and model transferability, highlighting the promise of initialization-based strategies for scalable and modular neural function approximation. The full code is made publicly available on Gitee.\footnotemark
\end{abstract}
\footnotetext{Gitee: \url{https://gitee.com/AIxinwen/nn4poly}.}

\maketitle


\section{Introduction}

Function approximation lies at the core of computational mathematics, computer science, and engineering, supporting both theoretical advances and practical applications. Many complex systems governed by differential or integral equations lack analytical solutions. In such cases, function approximation provides a tractable alternative by constructing numerical surrogates that enable efficient computation and analysis. These techniques play a fundamental role in various areas such as signal processing, numerical analysis, control theory, and uncertainty quantification \cite{1Powell}.

With the development of machine learning, function approximation has further evolved into a critical component of data-driven modeling. Modern approaches, especially those based on neural networks and support vector machines, have enabled remarkable advances in fields such as computer vision, speech recognition, and natural language processing \cite{2LeCun}. Among these, feedforward neural networks (FNNs) have attracted particular attention due to their universal approximation property: Any continuous function in a compact domain can be arbitrarily well approximated by an FNN with sufficient width \cite{3Hornik, 4Cybenko}.

Despite their expressive power, the practical use of neural networks for the approximation of functions faces several challenges.
First, networks employing classical activation functions (e.g., Sigmoid, Tanh) often suffer from vanishing or exploding gradients in training, particularly in deep architectures \cite{5Bengio}. Second, the highly non-convex optimization landscape of neural networks introduces numerous local minima, complicating convergence and stability \cite{6Glorot}. 
Third, neural networks are prone to overfitting, especially under limited or noisy data, which often leads to poor generalization \cite{8Srivastava}.
Finally, approximation accuracy is highly sensitive to the sampling strategy of training data: uniform sampling may fail to capture intricate behavior in sparse regions, whereas adaptive or hybrid strategies can provide better coverage. 

\begin{figure}[htb!]
    \centering
    \includegraphics[width=0.8\textwidth]{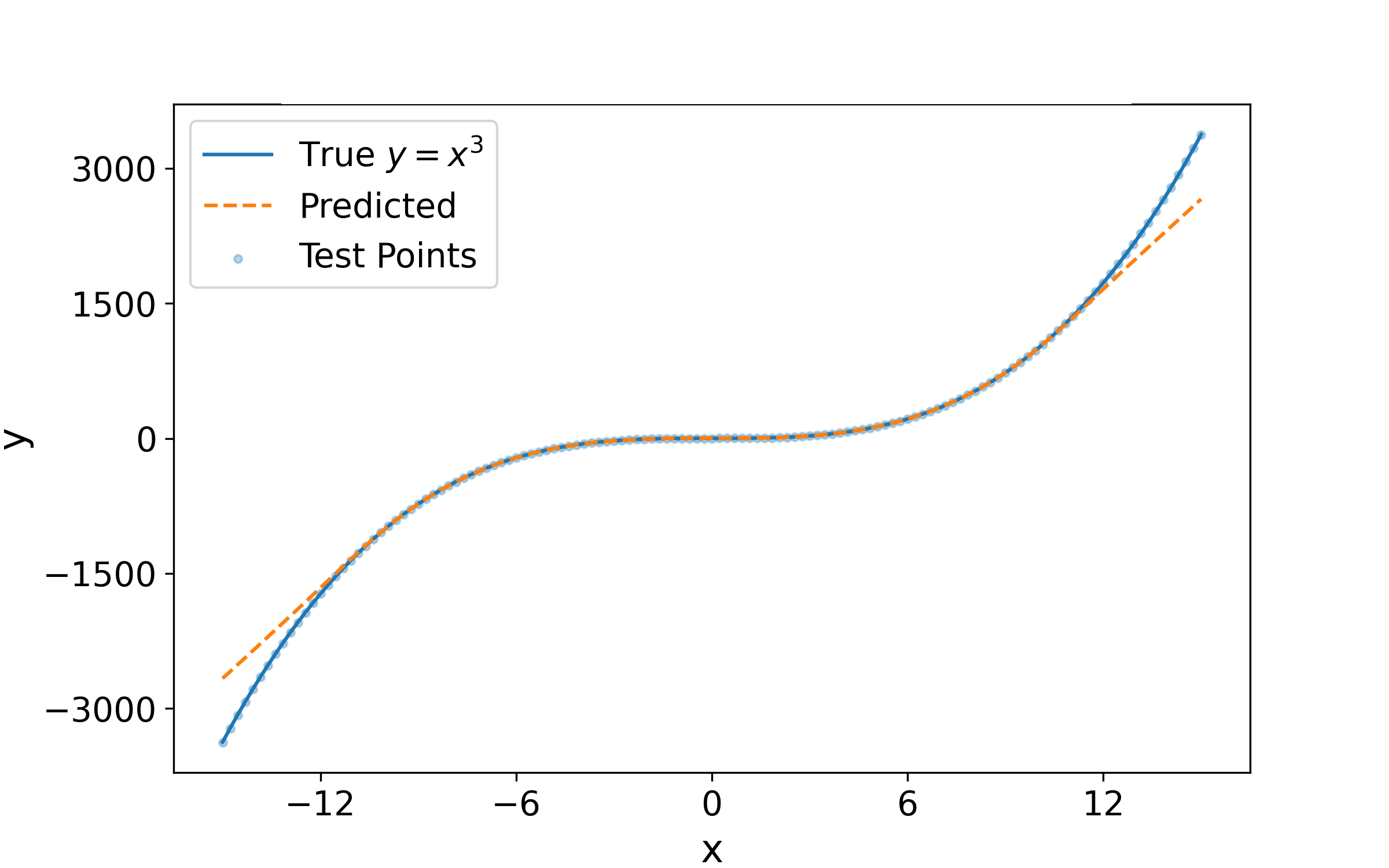}
    \caption{Extrapolation on $[-15, 15]$ from model trained on $[-10, 10]$.}
    \label{fig:1.2no_map}
\end{figure}

To illustrate the third challenge, we first introduce the mean squared error (MSE) and the coefficient of determination ($R^2$). The MSE is defined as
\[
\text{MSE} = \frac{1}{n} \sum_{i=1}^n \left( y_i - \hat{y}_i \right)^2,
\]
while the coefficient of determination is given by
\[
R^2 = 1 - \frac{\sum_{i=1}^n (y_i - \hat{y}_i)^2}{\sum_{i=1}^n (y_i - \bar{y})^2},
\]
with $y_i$ denoting the observed values, $\hat{y}_i$ the predicted values, and $\bar{y}$ the sample mean. The $R^2$ metric reflects the proportion of variance in the data explained by the predictions, with $R^2 = 1$ corresponding to perfect prediction (all data points lying exactly on the regression line).
Then, we examine a standard neural network trained to approximate the polynomial function \(f(x) = x^3\) on the interval \([-10,10]\). When evaluated on the larger interval \([-15,15]\), its extrapolation performance deteriorates sharply, as shown in Figure~\ref{fig:1.2no_map}. 
Quantitative results on \([-15,15]\) indicate that, although the network 
attains a coefficient of determination of $R^2 = 9.887 \times 10^{-1}$, 
indicating an overall satisfactory fit, the mean squared error reaches an unacceptable error \(3.8141\times 10^{6}\). This large MSE reflects substantial absolute deviations outside the training domain. The example underscores a fundamental challenge in neural approximation: the lack of reliable generalization beyond the training interval.

In addition to generalization, the initialization of network weights is another critical factor influencing neural approximation accuracy. To examine this effect, we conducted an experiment using a neural network with architecture \([2, 64, 64, 64, 1]\) to approximate the two dimensional function $f(x, y) = \sin(\pi x)\sin(4\pi y)$. 
The model was trained with different initialization strategies—Xavier, Kaiming, and Uniform—each evaluated under varying gain parameters.

\begin{figure}[htb!]
    \centering
    \includegraphics[width=0.8\textwidth]{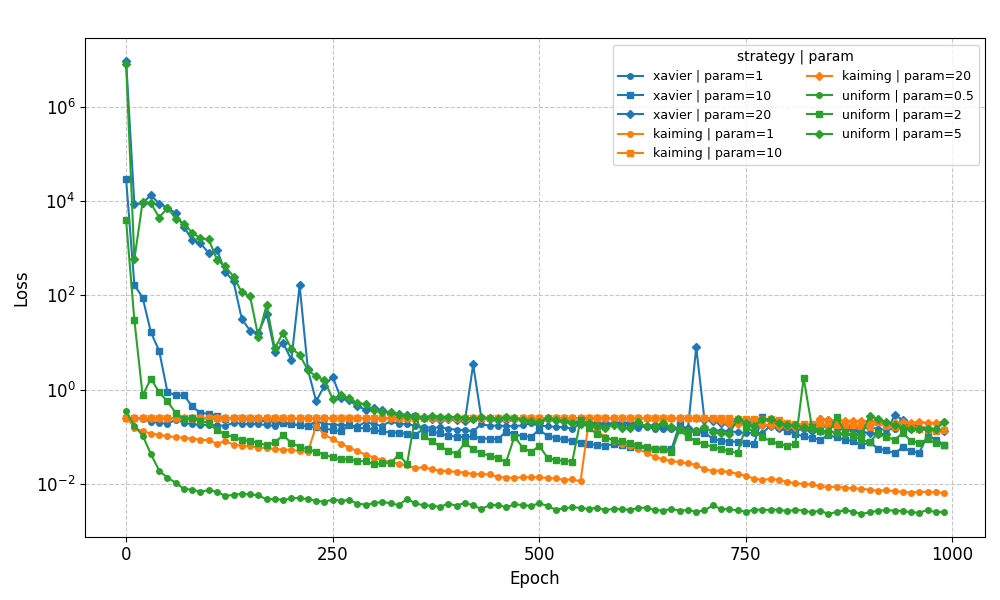}
    \caption{Training loss curves (log scale) under different initialization strategies.}
    \label{fig:init_loss_curve}
\end{figure}

As shown in \Cref{tab:init_sensitivity} and \Cref{fig:init_loss_curve}, the choice of initialization strategy has a significant impact on both the convergence of the network and the final approximation error. First, substantial differences are observed across initialization methods. For instance, Xavier initialization with a gain of $1$ yields a relative $L^2$ error of $7.27\times10^{-1}$, whereas Kaiming initialization with the same gain value produces a much smaller error of $1.55\times10^{-1}$. Second, even within the same initialization method, performance varies markedly with parameter settings. For example, Kaiming initialization results in relative $L^2$ errors of $1.58\times10^{-1}$, $7.53\times10^{-1}$, and $8.90\times10^{-1}$ for gains of $1$, $10$, and $20$, respectively, highlighting a strong sensitivity to the gain parameter. These findings indicate that initialization is not merely an implementation detail but a critical factor affecting both numerical stability and approximation accuracy.

\begin{table}[htb!]
    \centering
    \caption{Approximation errors for $ f(x, y) = \sin(\pi x)\sin(4\pi y) $ under different initialization strategies.}
    \label{tab:init_sensitivity}
    \begin{tabular}{|c|c|c|c|}
    \hline
    Strategy & Gain & Relative $L^2$ & MSE \\\hline
    Uniform & $0.5$ & $1.323735\times10^{-1}$   & $4.361765\times10^{-3}$ \\\hline
    Uniform & $2$   & $4.839527\times10^{-1}$   & $5.829962\times10^{-2}$ \\\hline
    Uniform & $5$   & $8.026694\times10^{-1}$   & $1.603738\times10^{-1}$ \\\hline
    Xavier  & $1$   & $7.266926\times10^{-1}$   & $1.314503\times10^{-1}$ \\\hline
    Xavier  & $10$  & $4.812306\times10^{-1}$   & $5.764563\times10^{-2}$ \\\hline
    Xavier  & $20$  & $7.010871\times10^{-1}$   & $1.223500\times10^{-1}$ \\\hline
    Kaiming & $1$   & $1.548526\times10^{-1}$   & $5.968940\times10^{-3}$ \\\hline
    Kaiming & $10$  & $7.528070\times10^{-1}$   & $1.410676\times10^{-1}$ \\\hline
    Kaiming & $20$  & $8.901272\times10^{-1}$   & $1.972260\times10^{-1}$ \\\hline
    \end{tabular}
\end{table}

To address the two main challenges discussed above---poor generalization and sensitivity to initialization---we propose a modular neural approximation framework that combines parameter reuse, domain normalization, and basis function pretraining. The central idea is to pretrain a collection of basis neural networks to approximate polynomial basis functions on a reference interval, such as $[-1,1]$. These pretrained modules then serve as reusable and composable components, which can be either directly assembled or fine-tuned to approximate more complex target functions.



Generalization to an arbitrary domain $[a,b] \subset \mathbb{R}$ is achieved through a domain normalization module that systematically maps inputs to the reference interval while preserving the structural properties of the function. This transformation enables seamless integration of pretrained basis networks into downstream approximation tasks, thereby enhancing transferability and supporting more reliable extrapolation.

The proposed method is systematically evaluated on a suite of benchmark functions in both one and two dimensions. We also analyze how key architectural factors—such as network depth, width, and activation function choice—influence approximation accuracy and generalization performance. Experimental results demonstrate notable improvements in efficiency, precision, and transferability, indicating the method’s potential as a scalable and general-purpose neural approximation approach.

The proposed method offers several key advantages:
\begin{itemize}
 \item 	Enhanced generalization: By consistently mapping inputs to a reference domain, the method mitigates extrapolation errors and enhances robustness.
 \item 	Efficiency through parameter reuse: Basis networks are trained once and reused across tasks, significantly accelerating convergence for downstream approximations.
 \item 	Modularity and extensibility: The framework naturally supports plug-and-play composition. Basis modules can be frozen as fixed feature extractors, or fine-tuned to improve expressivity. Combination coefficients can be computed analytically (e.g., via least squares) or learned end-to-end.
\end{itemize}


The remainder of this paper is organized as follows. Section 2 introduces the adopted function approximation method and model architecture, and explains how data mapping improves generalization, together with the strategy of pretraining and weights reuse for efficient network construction. Section 3 explains the influence of network depth, neuron count, and activation functions on the expressive power of basis functions, and presents inference results for one- and two-dimensional domains. Section 4 explains numerical experiments, applying the proposed initialization strategy to representative one- and two-dimensional functions with corresponding illustrations and error metrics. Section 5 concludes the paper and outlines future research directions.


\section{Reusable Weights Initialization for Efficient Function Approximation}

In function approximation, we are concerned with designing neural networks that can efficiently and accurately approximate a target function with enhanced generalization and training stability. We denote the space of continuous functions on $[a,b]$ by $\mathcal{C}([a,b])$. Let $f \in \mathcal{C}([a,b])$ be a continuous target function. We aim to construct an approximation $\mathcal{P}f$ using a set of pretrained basis functions $\hat{\varphi}_k$, such that:
\begin{equation}
\mathcal{P}f(x) = \sum_{k=0}^{K} \alpha_k \varphi_k(x),
\label{eq:probs}
\end{equation}
where:
\begin{itemize}
\item K denotes the number of basis functions in the least-squares procedure;
\item	$\mathcal{P}$ represents a projection operator induced by the basis functions;
\item	$\varphi_k(x) = \hat{\varphi}_k(\mathcal{T}^{-1}(x))$, where $\mathcal{T}$ is a domain transformation that maps the reference domain (e.g. $[-1,1]$ to $[a,b]$);
\item	The coefficients $\alpha_k$ may be determined using a least-squares approach.
\end{itemize}


Instead of training a new neural network from scratch for each target function---an approach that often suffers from slow convergence and limited generalization---we introduce a reusable weights initialization strategy built on pretrained basis network modules. The central idea is to first construct a library of compact neural networks, each trained to approximate a fundamental basis function (e.g., monomials) on a reference domain. These pretrained modules are then repurposed by transferring their learned weights to initialize the internal layers of a general-purpose function approximator.

\begin{figure}[htp!]
    \centering
    \includegraphics[width=0.75\linewidth]{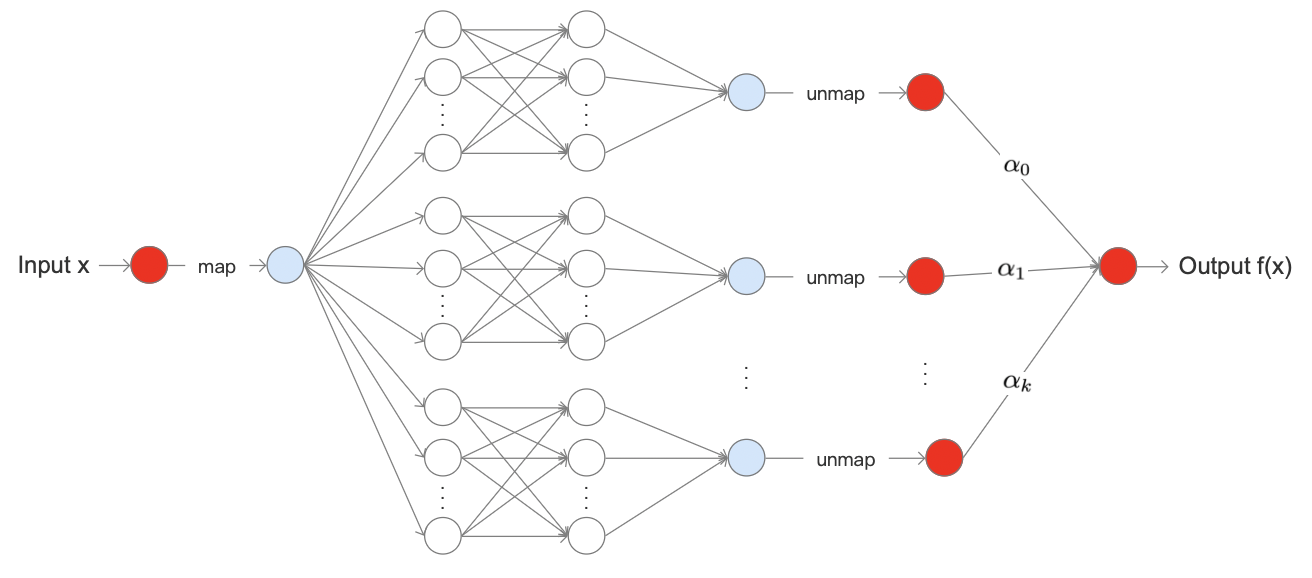}
    \caption{Workflow of the reusable weights initialization framework for function approximation. the left panel shows the pretraining of neural networks to approximate canonical basis functions on a reference domain. The learned weights are then reused to initialize the hidden layers of a general approximator (right panel), enabling faster convergence, improved generalization, and structural flexibility.}
    \label{fig:2.1two_stage}
\end{figure}

This initialization mechanism ensures that the approximator begins training from a meaningful initialization prior, leading to improved training efficiency, reduced overfitting, and better generalization over varying or large domains.
The overall workflow is illustrated in \ref{fig:2.1two_stage}, consisting of two conceptual phases:
\begin{itemize}
\item Construction of a library of pre-trained basis neural networks on a reference domain;
\item Deployment of these networks’ weights to initialize the hidden layers of the target approximation network, together with input transformations $\mathcal{T}$ to ensure domain compatibility.
\end{itemize}

To complement the schematic workflow illustrated in \Cref{fig:2.1two_stage}, \Cref{alg:reusable-init} formalizes the procedural details of the framework.

\begin{algorithm}[ht]
\caption{Reusable Weights Initialization Framework}
\label{alg:reusable-init}
\begin{algorithmic}[1]
\Require Target function $f(x)$, target domain $[a,b]$, reference domain $[-1,1]$, pretrained basis networks $\{\hat{\varphi}_0, \hat{\varphi}_1, \dots, \hat{\varphi}_M\}$, domain transformation $\mathcal{T}$
\Ensure Approximated function $\mathcal{P}f(x)$

\Statex \textbf{Offline: Library Construction}
\For{each basis function index $k = 0,1,\dots,M$}
    \State Train $\hat{\varphi}_k$ on reference domain $[-1,1]$ to approximate the $k$-th basis function
    \State Store trained parameters $\theta_k$ to formulate the library $V_{NN}$
\EndFor

\Statex \textbf{Online: Target Approximation}
\State Map input $x \in [a,b]$ to reference domain: $\hat x = \mathcal{T}(x)$
\State Initialize hidden layers with weights of pretrained models from $V_{NN}$ to obtain basis function values $\hat{\varphi}_k(\hat{x})$
\State Obtain $\hat{\varphi}_k(\mathcal{T}^{-1}(x))$ through unmapping basis function values to target domain using \eqref{unmap_form1} and \eqref{unmap_form2}
\State Construct target approximator $\mathcal{P}f(x) = \sum_{k=0}^{K} \alpha_k \hat{\varphi}_k(\mathcal{T}^{-1}(x))$ as the final output layer
\State Solve for coefficients $\{\alpha_k\}$ using least squares method
\State \Return $\mathcal{P}f(x)$
\end{algorithmic}
\end{algorithm}

\subsection{Pretraining basis neural networks}

To simplify the presentation, we fix the reference domain as $[-1,1]$. 
To approximate functions within this domain, we construct a set of basis neural 
networks that serve as accurate surrogates for the polynomial basis functions 
$\{\varphi_k\}_{k=0}^M$. Specifically, we define the neural representation space
\[
V_{\mathrm{NN}} = \left\{ \hat{\varphi}_k(\hat{x};\theta_k) \,\middle|\, k=0,\dots,M \right\},
\]
where $M$ denotes the number of functions in the basis library, and each 
$\hat{\varphi}_k \approx \varphi_k$ is a neural network approximation of the 
polynomial basis function, parameterized by trainable weights $\theta_k$. 
The basis networks are trained by minimizing the empirical loss
\[
L(\theta_k) = \frac{1}{N} \sum_{i=1}^{N} 
\left| \hat{\varphi}_k(\hat{x}_i; \theta_k) - \varphi_k(\hat{x}_i) \right|^2,
\]
where $\{\hat{x}_i\}_{i=1}^N$ are sampling points in the reference domain.

\begin{figure}[htp!]
    \centering
    \includegraphics[width=0.75\linewidth]{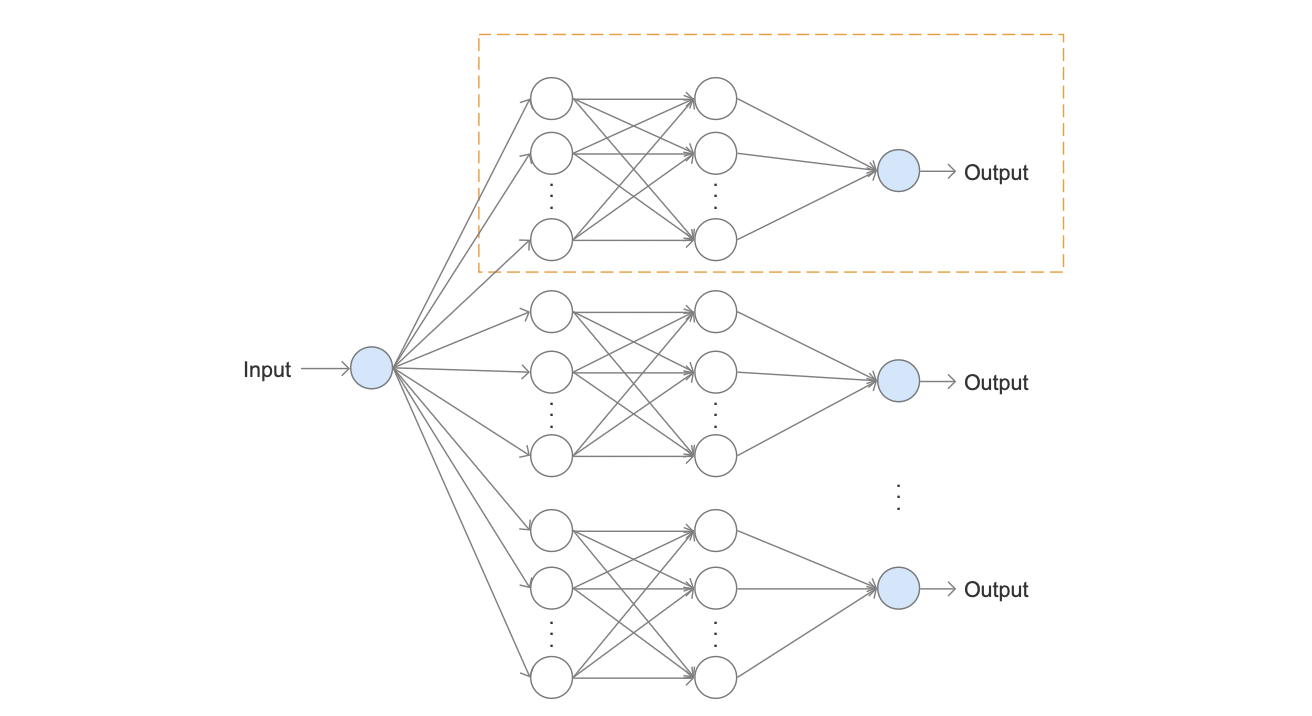}
    \caption{Modular construction of basis neural networks. Each module in orange box denotes a neural approximation to a specific basis function $\varphi_k$. These modules can be reused and composed for downstream tasks.}\label{fig:2.1two_stage_1}
\end{figure}

Once trained, the resulting basis neural networks form a reusable library of basis function modules. As illustrated in Figure~\ref{fig:2.1two_stage_1}, each module can be interpreted as a functional building block, capable of being composed or embedded into more complex neural architectures. The weights of these modules may either be fixed to preserve their pre-learned behavior or fine-tuned to accommodate task-specific adaptations.



To efficiently approximate structured function families such as the polynomial basis functions $\varphi_k(x) = x^k$, $k=0,1,\ldots$, we employ a progressive initialization strategy. Rather than training each network independently from scratch, the parameters obtained from the model trained on $\varphi_{k-1}(x)$ are used to initialize the training for $\varphi_k(x)$. This sequential reuse of parameters provides informed initialization, thereby accelerating convergence and enhancing training stability. The procedure is outlined in \Cref{alg:progressive-init}.

\begin{algorithm}[ht]
\caption{Progressive Initialization}
\label{alg:progressive-init}
\begin{algorithmic}[1]
\Require Function family $\{\varphi_k(x)\}$, maximum degree $M$.
\Ensure Trained models $\{\hat{\varphi}_0, \hat{\varphi}_1, \dots, \hat{\varphi}_M\}$ with parameters $\{\theta_0, \theta_1, \dots, \theta_M\}$
\State Initialize $\theta_0^{(0)}$ randomly
\State Train model $\hat{\varphi}_0(x; \theta_0) \approx \varphi_0(x) = x^0$ with $\theta_0^{(0)}$
\For{$k = 1$ to $M$}
    \State $\theta_k^{(0)} \gets \theta_{k-1}$
    \State Initialize model $\hat{\varphi}_k$ with $\theta_k^{(0)}$
    \State Train $\hat{\varphi}_k(x; \theta_k) \approx \varphi_k(x) = x^k$
\EndFor
\State \Return $\{\theta_0, \theta_1, \dots, \theta_M\}$
\end{algorithmic}
\end{algorithm}

\begin{remark}\label{rem2e1}
This initialization strategy leverages the inductive bias between adjacent basis functions, yielding substantial improvements in training efficiency. Similar inductive biases have been validated in studies on transfer learning \cite{Houlsby2019, Li2018}.
\end{remark}

\subsection{Constructing approximation via basis neural networks and data mapping}
To extend the applicability of pretrained basis neural networks from the reference domain $[-1,1]$ to general intervals $[a,b] \subset \mathbb{R}$, we introduce a data mapping strategy that enables consistent reuse of learned representations to general domains. 

We begin by defining a bidirectional mapping between the computational interval $[a,b]$ and the reference interval $[-1,1]$:
$$ \begin{cases}\hat x=T(x;[a,b])&\text{(Forward mapping)},\\
x=T^{-1}(\hat x;[-1,1])&\text{(Inverse mapping)}.
\end{cases} 
$$
Here, the tailored mapping is realized via a logarithmic scaling:
$$ T(x)=\frac{x}{10^s},\quad s(x) = \Big\lceil \log_{10}\big(\lfloor |x|+1 \rfloor \big) \Big\rceil,
$$
where the exponent $s$ is computed only during the forward mapping and remains fixed for each input $x$.

\begin{figure}[htb!]
    \centering
    \includegraphics[width=0.75\linewidth]{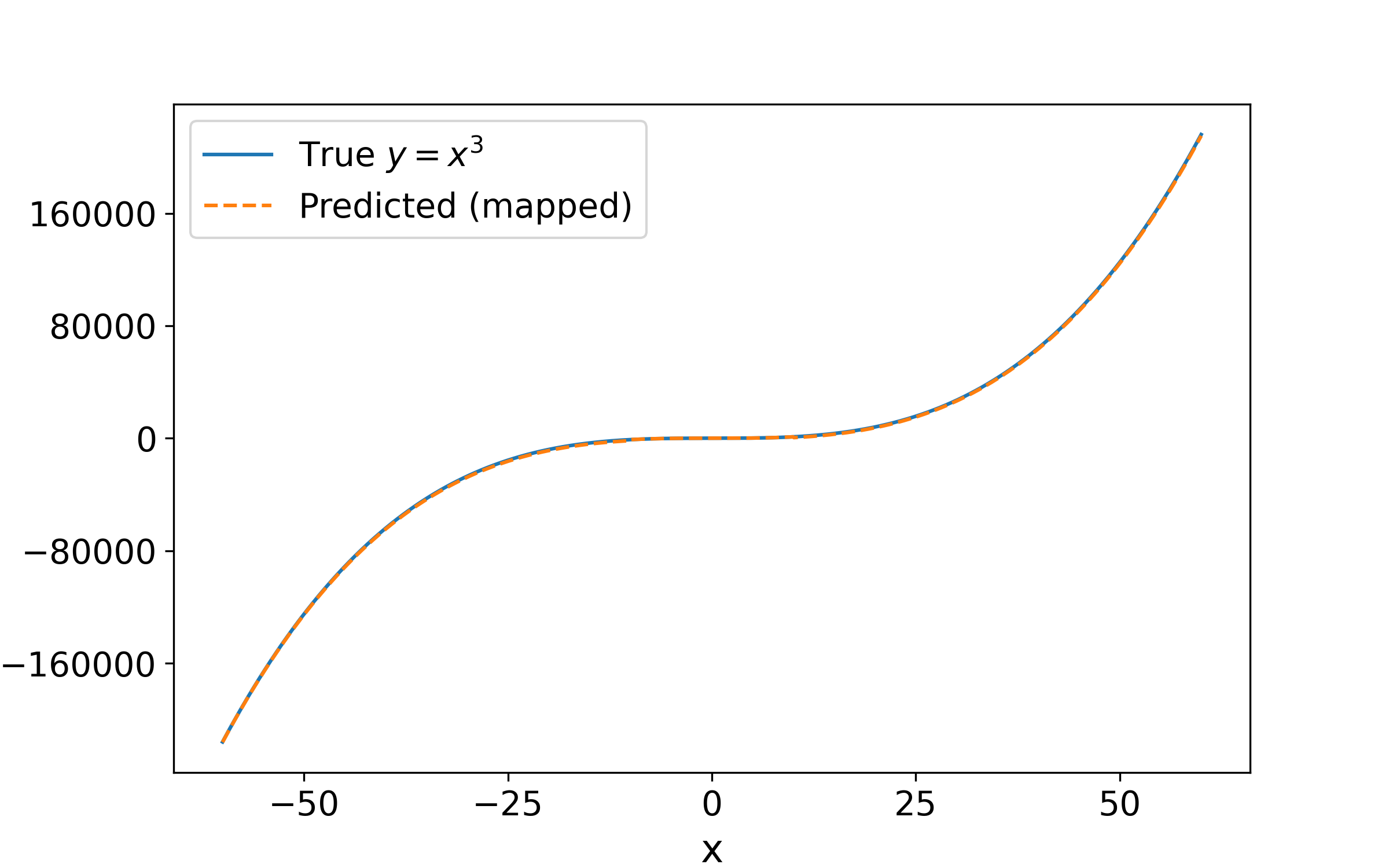}
    \caption{Model inference of $y = x^3$ over $[-60,60]$ after inverse mapping.}
    \label{fig:2.3.4generation}
\end{figure}

To illustrate the improved generalization ability of neural networks with this mapping, we revisit the approximation of the function \(y = x^3\) over a much larger interval, \(x \in [-60,60]\), compared to \Cref{fig:1.2no_map}. After applying the transformation \(x \mapsto \hat{x}\), a neural network originally trained on \([-1,1]\) is repurposed to approximate the transformed function \(\hat{y} = \hat{x}^3\). The inference results on \(x \in [-60,60]\) are presented in \Cref{fig:2.3.4generation}, where the inverse mapping restores predictions in the original domain. Quantitative evaluation yields a mean squared error (MSE) of \(2.87304 \times 10^{1}\) and a coefficient of determination of \(R^2 = 9.998 \times 10^{-1}\).

Once the neural approximation $\hat{\varphi}(\hat{x})$ has been trained on the reference interval, the corresponding functions $\varphi(x)$ on any computational interval $[a,b]$ can be recovered through inverse mapping. Specifically, for a basis function of the form $\varphi(x) = x^k$, the neural approximation satisfies
\begin{equation}
\hat{\varphi}(\hat{x}) = \hat{x}^k, \quad \text{where } \hat{x} = \frac{x}{10^s}.
\label{unmap_form1}
\end{equation}
The corresponding recovery formula becomes:
\begin{equation}
\varphi(x) = 10^{ks} \cdot \hat{\varphi}(\hat{x}).
\label{unmap_form2}
\end{equation}
where $s = \Big\lceil \log_{10}\big(\lfloor |x|+1 \rfloor \big) \Big\rceil$ again encodes the original scaling.


This weights-reuse framework enables neural models trained on the reference domain to be seamlessly extended to arbitrary real-valued intervals without retraining from scratch. In downstream applications—such as functional regression or spectral projection—the reused basis networks provide efficient parameter initialization and retain generalization capability. For a target function $f \in \mathcal{C}([a,b])$, one can determine the coefficients $\alpha_k$ by projecting $f$ onto the mapped basis $\varphi_k(x)$, as described in \eqref{eq:probs}.


\section{Optimal basis neural networks}
In function approximation problems, polynomials are commonly employed as basis functions. However, different polynomial bases exhibit varying levels of complexity and nonlinearity. In practical applications, hardware constraints and computational resources further necessitate careful selection of the neural network architecture.

The problem can be reformulated as follows: Given GPU/CPU resource constraints and limits on training time, our goal is to optimize key neural network hyperparameters—such as depth, width, and activation functions—to balance accuracy with efficient inference \cite{25Shen2021, 28Lu2017}. Specifically, we investigate:
\begin{itemize}
    \item Network depth (number of layers): The number of layers in a neural network determines the model's depth. Excessively deep networks may lead to prolonged training times and an increased risk of overfitting, yet they are often necessary when approximating complex functions. Common choices for the number of layers typically include $1, 2, 4$, or $8$ layers.
    \item Width (neurons per layer): The number of neurons per layer affects the network's representational capacity. More neurons enable the approximation of more complex functions, but they also increase computational and storage costs. Typically, the number of neurons per layer ranges from $16$ to $1024$.
    \item Activation functions: Choosing appropriate activation functions can enhance the network's nonlinear representation capability while mitigating the risks of gradient vanishing or explosion. Common activation functions available for selection include ReLU, Tanh, GELU, Mish, and ELU.
\end{itemize}

To maximize approximation accuracy with limited resources. The search space for architecture selection depends on target function complexity, dataset size and available hardware resources.

\subsection{Network Architecture}    
Without loss of generality, we take the function $y = x^3$ as an example to examine the influence of the number of neurons and the number of hidden layers in the network on the approximation effect.
Specifically, we set 
\begin{itemize}
    \item Training Data: Sampled in $[-1, 1]$.
    \item Test Data: Sampled in $[-20, 20]$ (evaluating generalization).
\end{itemize}

We first consider the single hidden-layer neural network, and the corresponding results are reported in Table \ref{tab:3onelayer}. we can see clearly: i) as the number of neurons increases $ (8\rightarrow 32) $, the MSE decreases from $ 8.57\times 10^{-7} $ to $ 4.23\times 10^{-8} $, demonstrating that expanding model capacity enhances fitting accuracy on training data; ii) with $64$ neurons, the MSE exhibits minor fluctuations and the smallest MSE is achieved with $256$ neurons.

In terms of network depth, the results in Table \ref{tab:3manylayersamecell8}-\ref{tab:3manylayersamecell128} show that increasing the number of hidden layers can significantly improve approximation accuracy when each layer is relatively narrow. For example, extending a shallow architecture like $[1, 8, 1]$ to $[1, 8, 8, 1]$ or $[1, 8, 8, 8, 8, 1]$ can reduce the mean squared error (MSE) by an order of magnitude. However, when the network is already wide (e.g., $128$ neurons per layer), further increasing the depth may instead lead to performance degradation, which may be attributed to overfitting or optimization difficulties. Therefore, a depth of $1$ to $4$ layers provides a good balance between expressiveness and training efficiency.

\begin{table}
    \centering
    \caption{Approximation $ y=x^3 $ using single hidden layer neural networks with progressively increasing neurons.}\label{tab:3onelayer}
    \begin{tabular}{|l|l|l|}
    \hline
    Net & MSE&  $R^2$\\\hline
    [1, 8, 1]  &  $8.57\times 10^{-7}$ &    $9.99989\times 10^{-1}$\\\hline
    [1, 16, 1]  &  $5.84\times 10^{-8}$ &    $9.99999\times 10^{-1}$\\\hline
    [1, 32, 1] &  $4.23\times 10^{-8}$ &   $9.99999\times 10^{-1}$\\\hline
    [1, 64, 1] &  $1.15\times 10^{-7}$ &    $9.99999\times 10^{-1}$\\\hline
    [1, 128, 1] & $ 8.08\times 10^{-8}$ &  $ 9.99999\times 10^{-1}$\\\hline
    [1, 256, 1] &  $8.76\times 10^{-9}$ &   $ 1.0 $\\\hline
    \end{tabular}
\end{table}

\begin{table}
    \centering
    \caption{Neural networks with increasing hidden layers (fixed $8$ neurons per layer) for approximating $ y=x^3 $.}\label{tab:3manylayersamecell8}
    \begin{tabular}{|l|l|l|}
    \hline
    Net & MSE&  $R^2$\\\hline
    [1, 8, 1]  &  $8.57\times 10^{-7}$ &   $9.99989\times 10^{-1}$\\\hline
    [1, 8, 8, 1]  &  $5.39\times 10^{-8}$ &    $9.99999\times 10^{-1}$\\\hline
    [1, 8, 8, 8, 8, 1]  &  $4.48\times 10^{-8}$ &    $9.99999\times 10^{-1}$\\\hline
    \end{tabular}
\end{table}

\begin{table}
    \centering
    \caption{Approximating $ y=x^3 $ using neural networks with increasing hidden layers and fixed width ($128$ neurons per hidden layer).}\label{tab:3manylayersamecell128}
    \begin{tabular}{|l|l|l|}
    \hline
    Net & MSE & $R^2$\\\hline
    [1, 128, 1] &  $8.08\times 10^{-8}$ &    $9.99999\times 10^{-1}$\\\hline
    [1, 128, 128, 1] &  $1.5\times 10^{-8}$ &    $1.0$\\\hline
    [1, 128, 128, 128, 128, 1] &  $3.16\times 10^{-7}$ &  $9.99996\times 10^{-1}$\\\hline
    \end{tabular}
\end{table} 



Next, we examine the impact of network depth on function approximation performance.
As shown in Table \ref{tab:3manylayersamecell8}, when fixing the width at $8$ neurons per hidden layer, increasing the number of layers progressively enhances the network’s expressive power via nonlinear compositional operations, thereby compensating for the limited capacity of single-layer architectures—an effect reminiscent of the depth advantage in residual networks \cite{10He}. In contrast, for wider networks with $128$ neurons per hidden layer (Table \ref{tab:3manylayersamecell128}), the best performance is achieved with two hidden layers, while further increasing depth to four layers leads to performance degradation. This underscores the necessity of balanced width–depth scaling, in line with the compound scaling principle of EfficientNet \cite{11Tan}.


For a comprehensive evaluation, we systematically varied both the number of neurons and the depth of the network while keeping the data sampling protocol fixed: training samples were drawn from $[-1,1]$, whereas the test set was extended to $[-20,20]$ to rigorously assess generalization beyond the training distribution. This experimental design enables a controlled numerical investigation of the interplay between network width and depth.

\begin{table}
    \centering
    \caption{Neural networks with varied hidden layers and node counts for approximating $ y=x^3 $.}\label{tab:3manylayermanycells}
    \begin{tabular}{|l|l|l|}
    \hline
    Net & MSE & $R^2$\\\hline
    [1, 8,1]  & $8.57\times 10^{-7}$ &  $9.99989\times 10^{-1}$\\\hline
    [1, 8, 8, 1]  & $5.39\times 10^{-8}$ &  $9.99999\times 10^{-1}$\\\hline
    [1, 8, 16, 1]  & $3.34\times 10^{-7}$ & $9.99996\times 10^{-1}$\\\hline
    [1, 8, 32, 1]  & $1.36\times 10^{-7}$ & $9.99998\times 10^{-1}$\\\hline
    [1, 8, 64, 1]  & $1.81\times 10^{-7}$ & $9.99998\times 10^{-1}$\\\hline
    [1, 8, 128, 1]  & $7.26\times 10^{-7}$ & $9.99991\times 10^{-1}$\\\hline
    [1, 8, 256, 1]  & $9.63\times 10^{-9}$ & $ 1.0$\\\hline
    [1, 8, 8, 8, 8, 1]  & $4.48\times 10^{-8}$ & $9.99999\times 10^{-1}$\\\hline
    [1, 8, 16, 32, 64, 1]  & $6.37\times 10^{-9}$ & $ 1.0$\\\hline
    [1, 8, 16, 32, 128, 1]  & $1.3\times 10^{-8}$ & $ 1.0$\\\hline
    [1, 8, 16, 32, 256, 1]  & $8.74\times 10^{-9}$ & $ 1.0$\\\hline
    [1, 8, 16, 64, 128, 1]  & $3.9\times 10^{-8}$ & $ 1.0$\\\hline
    [1, 8, 16, 64, 256, 1]  & $5.06\times 10^{-9}$ & $ 1.0$\\\hline
    [1, 8, 16, 128, 256, 1]  & $1.74\times 10^{-8}$ & $ 1.0$\\\hline
    [1, 8, 32, 64, 128, 1]  & $2.94\times 10^{-9}$ & $ 1.0$\\\hline
    [1, 8, 32, 64, 256, 1]  & $2.98\times 10^{-8}$ & $ 1.0$\\\hline
    [1, 8, 32, 128, 256, 1]  & $3.8\times 10^{-8}$ & $ 1.0$\\\hline
    [1, 8, 64, 128, 256, 1]  & $2.1\times 10^{-9}$ & $ 1.0$\\\hline
    \end{tabular}
\end{table}  


The results in Table \ref{tab:3manylayermanycells} reveal marked performance differences across network architectures. A single hidden layer network with configuration $[1, 8, 1]$ attains an MSE of $8.57\times 10^{-7}$, substantially higher than that of deeper models, highlighting the critical role of depth in enhancing representational capacity. Expanding to a two-hidden-layer architecture $[1, 8, 8, 1]$ lowers the MSE to $5.39\times 10^{-8}$, showing that additional depth—even without increasing width—can significantly strengthen nonlinear approximation capabilities.


Further performance improvements are observed in progressively wider architectures: the $[1, 8, 16, 32, 64, 1]$ network reduces the MSE to $6.37\times 10^{-9}$, while the $[1, 8, 64, 128, 256, 1]$ configuration achieves an impressive MSE of $2.1\times 10^{-9}$. These results demonstrate that gradually increasing the number of neurons across layers effectively expands model capacity while maintaining training stability. Such architectures enable hierarchical feature learning, where lower layers capture low-frequency patterns and higher layers extract fine-grained details, consistent with established hierarchical representation theory \cite{12Bengio}. The consistent improvements across these systematically scaled architectures provide strong empirical evidence for the joint importance of depth and progressive width scaling in neural network design.

In terms of network width, increasing the number of neurons per layer also enhances representational power and reduces fitting error, particularly in shallow networks. For instance, increasing the width from $8$ to $256$ neurons in a single hidden layer architecture significantly reduces the MSE. Notably, good performance is also achieved by architectures that combine depth with progressively increasing layer widths, such as $[1, 8, 16, 32, 64, 1]$ or $[1, 8, 64, 128, 256, 1]$. These structures not only achieve extremely low training error but also exhibit strong generalization. From the perspective of accuracy, both depth and width are beneficial. However, considering model complexity and computational performance, a single hidden layer network with sufficient width is a practical and effective choice.


Based on these experimental findings, and considering the dual requirements of accuracy and computational efficiency for basis function approximation within the training domain, all subsequent experiments employ a standardized architecture with a single hidden layer containing $1024$ neurons. This configuration provides sufficient expressive power to capture the complexity of polynomial functions while remaining computationally tractable across different basis functions and training settings. In addition, the domain mapping module introduced in this work ensures robust out-of-domain generalization of the learned basis functions.

\subsection{Activation Functions}
The activation function plays a pivotal role in both the design and training of neural networks. These nonlinear transformations not only determine a model's expressive power but also directly impact computational efficiency during training, particularly when processing large-scale datasets or complex architectures. With the advancement of deep learning, an expanding repertoire of activation functions has emerged, each demonstrating distinct effectiveness and efficiency across varied tasks.

Under controlled experimental conditions (macOS 15.3.1, Python 3.11.9), systematic measurements of forward and backward propagation times were conducted over $ N $ iterations to evaluate computational efficiency for the following activation functions
\begin{itemize}
\item ReLU\cite{19Glorot}: \[ReLU(x)=\max(0,x)\]

\item Sigmoid\cite{20Rumelhart}: \[S(x)=\frac{1}{1+e^{-x}}=\frac{e^x}{1+e^x};\]

\item Tanh\cite{16LeCun}: \[T(x)=\frac{e^x-e^{-x}}{e^x+e^{-x}};\]

\item Mish\cite{14Misra}: \[M(x)=x\cdot \tan(\ln(1+e^x));\]

\item GELU\cite{15Hendrycks}: \[GELU(x)=0.5x\left(1+\tanh\left(\sqrt{2}{\pi}(x+0.047715x^2)\right)\right);\]

\item SELU\cite{18Klambauer}: \[SELU(x)=\begin{cases}\lambda x & x>0,\\ \lambda\alpha (e^x-1)& x\leq 0;\end{cases}\]

\item CELU\cite{17Barron}: \[CELU(x)=\max(0,x) + \min(0,\alpha(e^{x/a}-1)).\]
\end{itemize}

\begin{figure}[htp!]
    \centering
    \includegraphics[width=0.75\linewidth]{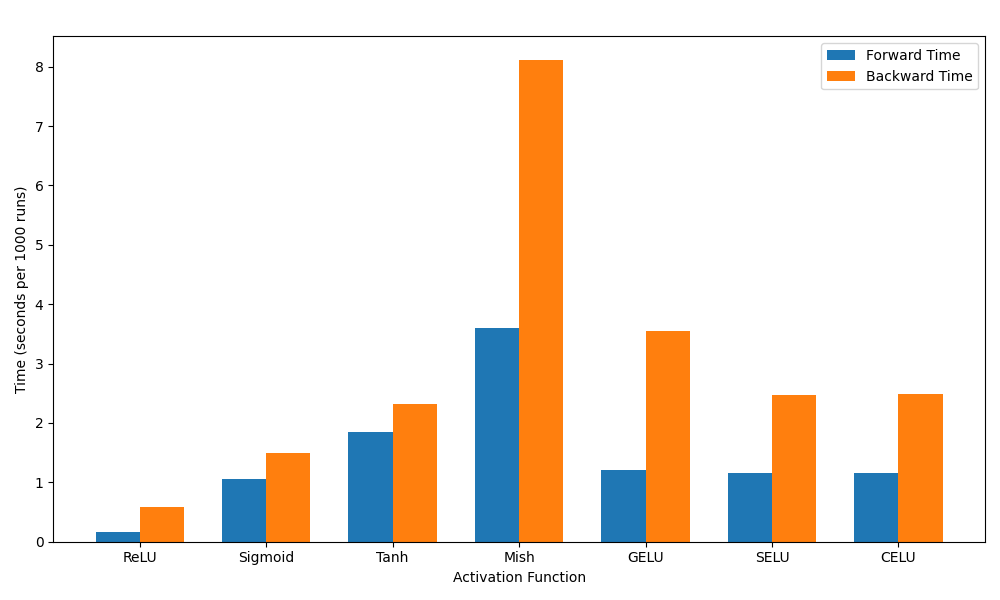}
    \caption{Comparative time cost of $1000$ computations across activation functions.}\label{fig:activation_efficiency}
\end{figure}

As evidenced in Figure \ref{fig:activation_efficiency}, ReLU demonstrates superior computational speed owing to its simple thresholding operation that eliminates the need for exponential or complex mathematical computations. In contrast, Sigmoid and Tanh functions exhibit significantly longer propagation times - approximately $2-3$ times slower than ReLU \cite{13Nair} - due to their inherent dependence on exponential function evaluations.

The most computationally intensive activations prove to be Mish and GELU, where compound nonlinearities and approximation requirements substantially increase processing overhead. Specifically, GELU's reliance on error function approximations and Mish's composite nonlinear operations result in backward propagation times that are $4-5$ times greater than those of ReLU. These empirical findings quantitatively validate the critical trade-off between activation function expressivity and computational efficiency in deep learning systems.

Similarly, using polynomial fitting of $ y=x^3 $ as an example, we present evaluation metrics (Table \ref{tab:3activationerror}) for networks employing different activation functions while maintaining fixed architecture and training hyperparameters:

\begin{table}
    \centering
    \caption{Comparative analysis of generalization error metrics using different activation functions in neural networks approximating $ y=x^3 $.}\label{tab:3activationerror}
    \begin{tabular}{|l|l|l|}
    \hline
    Function & MSE  & $R^2$\\\hline
    GELU & $1.40046575\times 10^{-6}$ &  $9.99994499\times 10^{-1}$\\\hline
    CELU & $1.60336242\times 10^{-6}$ &  $9.99993702\times 10^{-1}$\\\hline
    SELU & $1.71072879\times 10^{-5}$ &  $9.99932807\times 10^{-1}$\\\hline
    Mish & $8.07827476\times 10^{-8}$ &  $9.99999683\times 10^{-1}$\\\hline
    Tanh & $8.50998640\times 10^{-6}$ &  $9.99966575\times 10^{-1}$\\\hline
    ReLU & $9.86716373\times 10^{-7}$ &  $9.99996124\times 10^{-1}$\\\hline
    Sigmoid & $3.42593557\times 10^{-4}$ &  $9.98654388\times 10^{-1}$\\\hline
    \end{tabular}
\end{table} 

As fundamental components of neural networks, activation functions critically determine a model's nonlinear representational capacity. Empirical studies reveal substantial performance variations among different activation functions across diverse tasks.

The Mish activation function effectively combines ReLU's nonlinear characteristics with Tanh's smoothness, while its distinctive non-monotonic property enables superior capture of higher-order nonlinear patterns in polynomial approximation tasks. While ReLU offers notable advantages in training efficiency and mitigation of vanishing gradient problems, its application in certain function fitting scenarios may lead to suboptimal local minima or real-domain discontinuities, potentially compromising approximation accuracy. GELU employs Gaussian error function-based smoothing of negative values to maintain nonlinearity while enhancing gradient stability, whereas CELU utilizes exponential transformations for negative inputs at the cost of requiring additional parameter optimization (e.g. coefficient tuning). 

In summary, our comprehensive evaluation of commonly used activation functions reveals a clear trade-off between computational efficiency and approximation accuracy. While ReLU remains the fastest option due to its minimalist structure, it lacks the expressive smoothness required in higher-order function approximation. Mish demonstrates superior fitting performance but suffers from prohibitively high computational cost, particularly during backpropagation.

Considering both performance and efficiency, GELU emerges as the most balanced choice. It offers excellent generalization capability—comparable to Mish—with significantly lower computational overhead, especially when compared to other smooth, non-monotonic activations. Its incorporation of the Gaussian error function contributes to improved gradient stability and smoother approximation across nonlinear regions, making it well-suited for the function approximation tasks considered in this study.

Therefore, in all subsequent experiments and network implementations throughout this work, GELU is adopted as the default activation function, providing an optimal compromise between accuracy and computational cost.

\subsection{Performance of basis neural networks}   

\begin{figure}[htb!]
    \centering
    \begin{minipage}[b]{0.3\textwidth}
        \includegraphics[width=\textwidth]{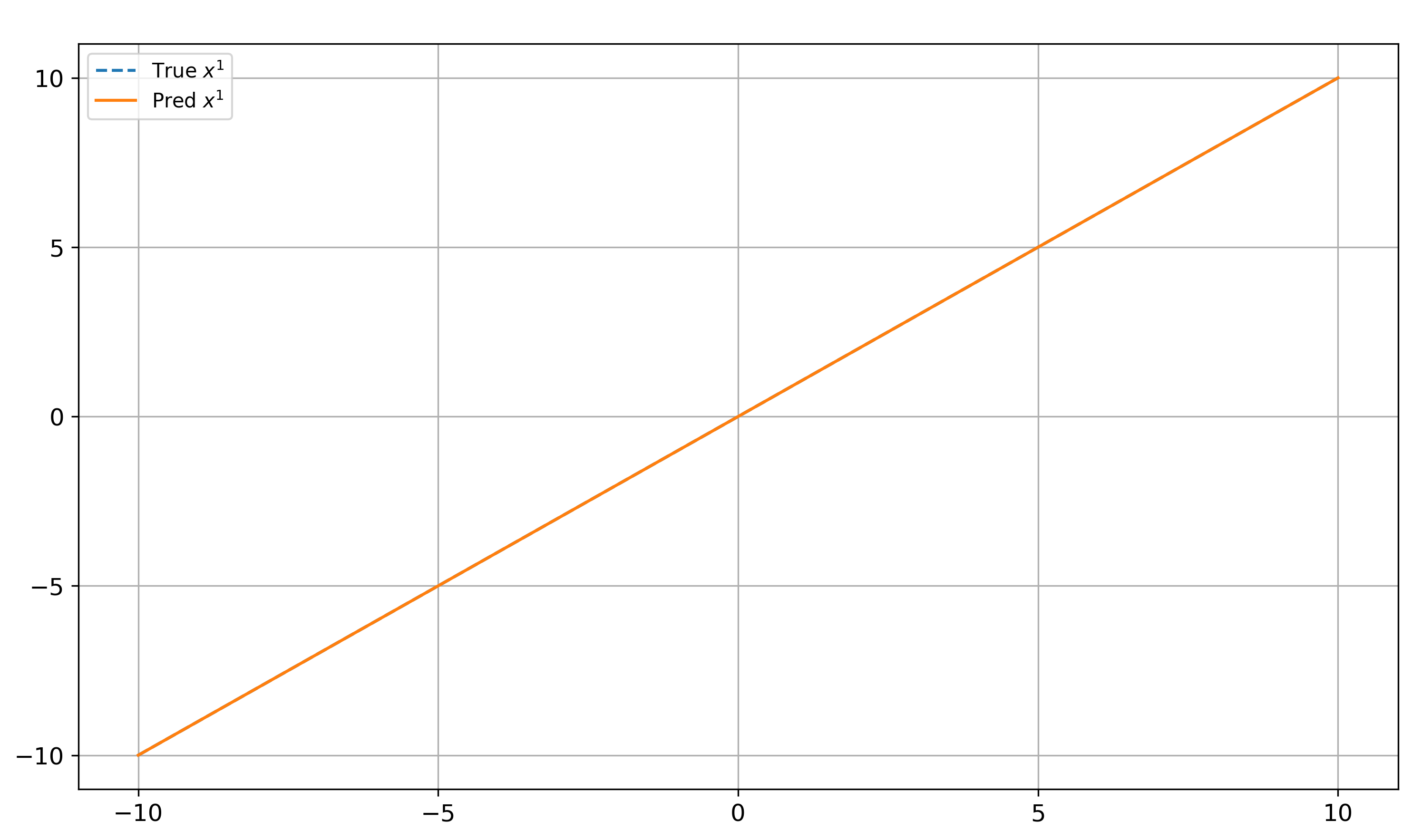}
        \caption*{$f(x) = x^1$}
    \end{minipage}
    \hfill
    \begin{minipage}[b]{0.3\textwidth}
        \includegraphics[width=\textwidth]{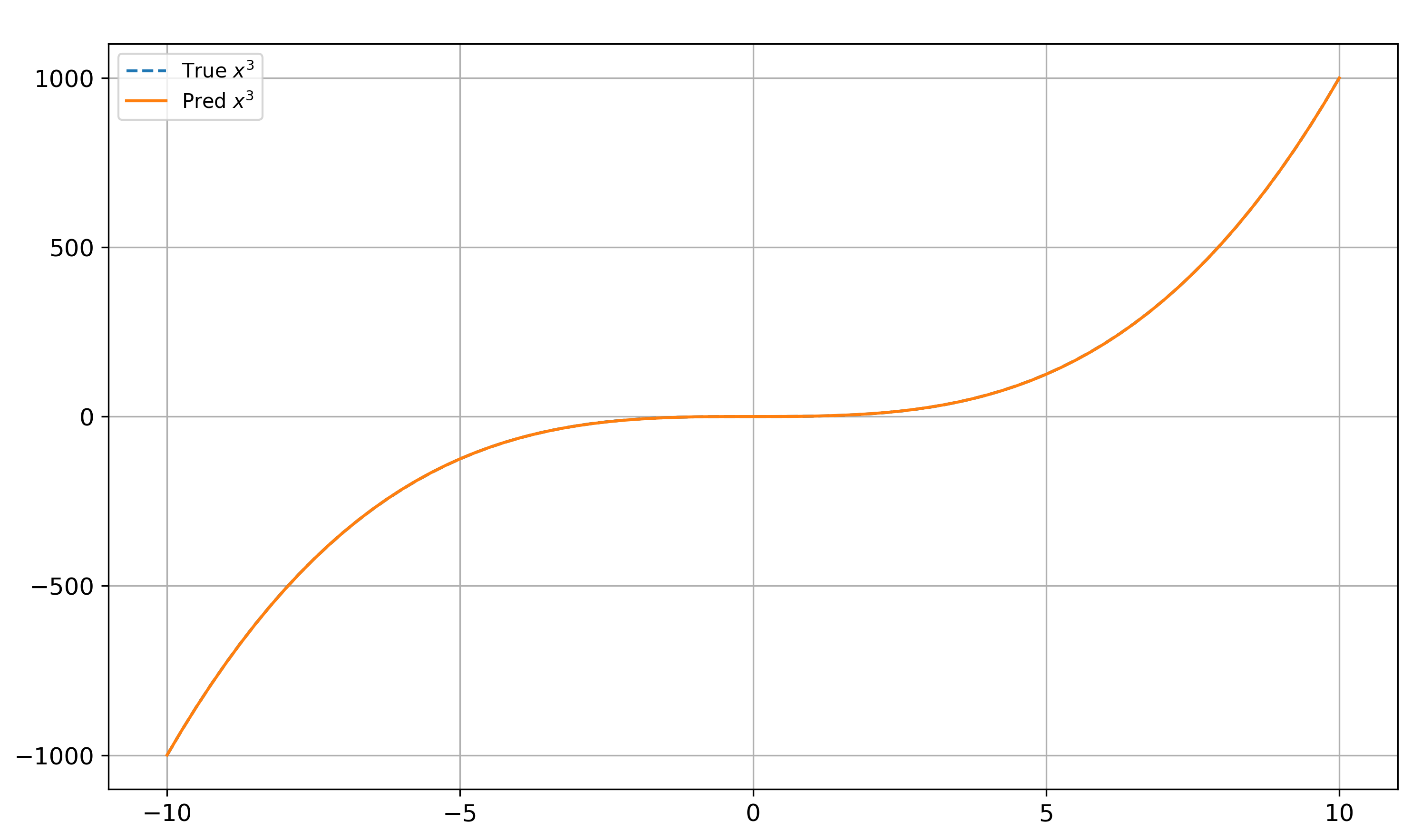}
        \caption*{$f(x) = x^3$}
    \end{minipage}   
    \hfill
    \begin{minipage}[b]{0.3\textwidth}
        \includegraphics[width=\textwidth]{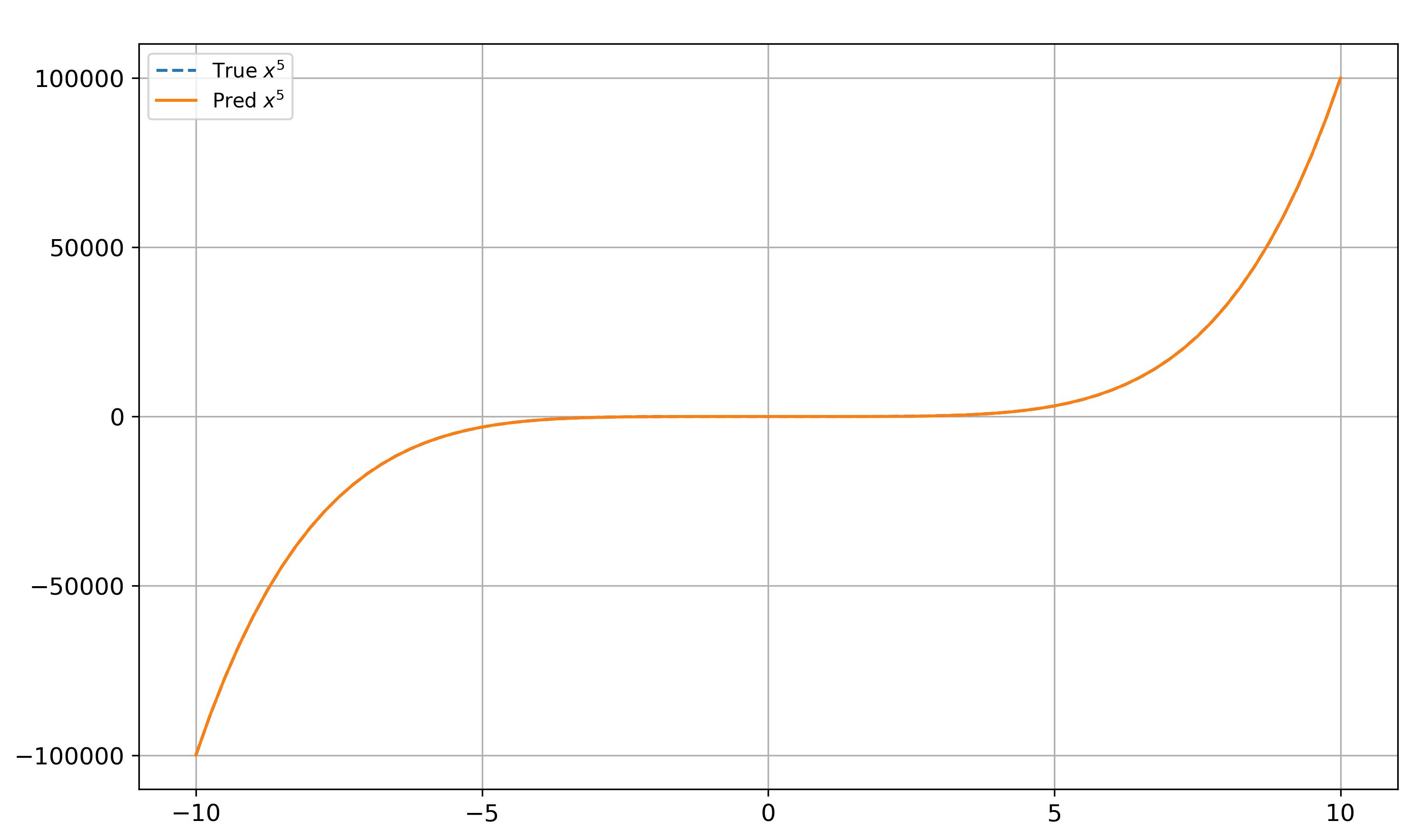}
        \caption*{$f(x) = x^5$}
    \end{minipage}
   
    \vspace{0.5cm} 
  
     \begin{minipage}[b]{0.3\textwidth}
         \includegraphics[width=\textwidth]{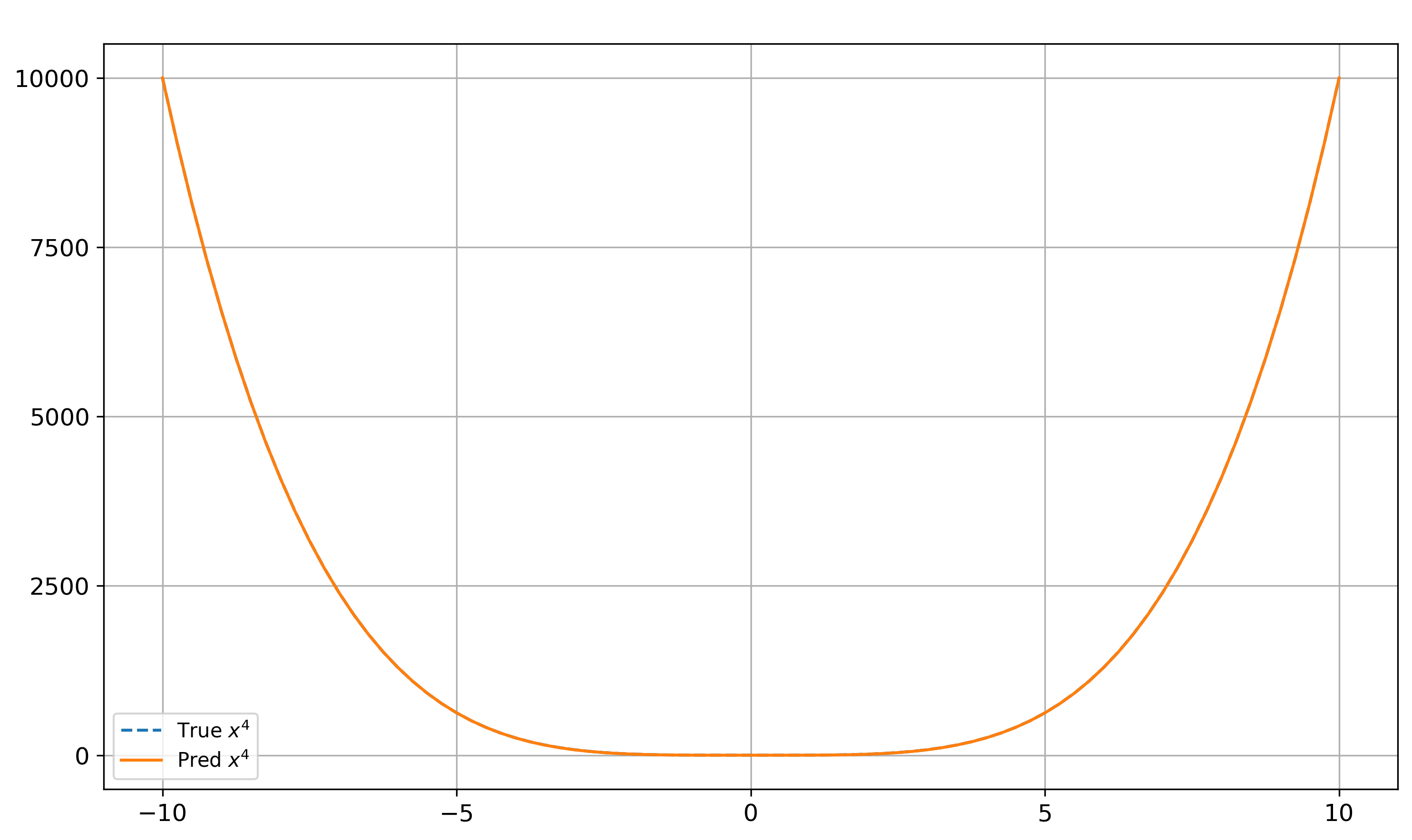}
         \caption*{$f(x) = x^4$}
     \end{minipage}
     \hfill
     \begin{minipage}[b]{0.3\textwidth}
         \includegraphics[width=\textwidth]{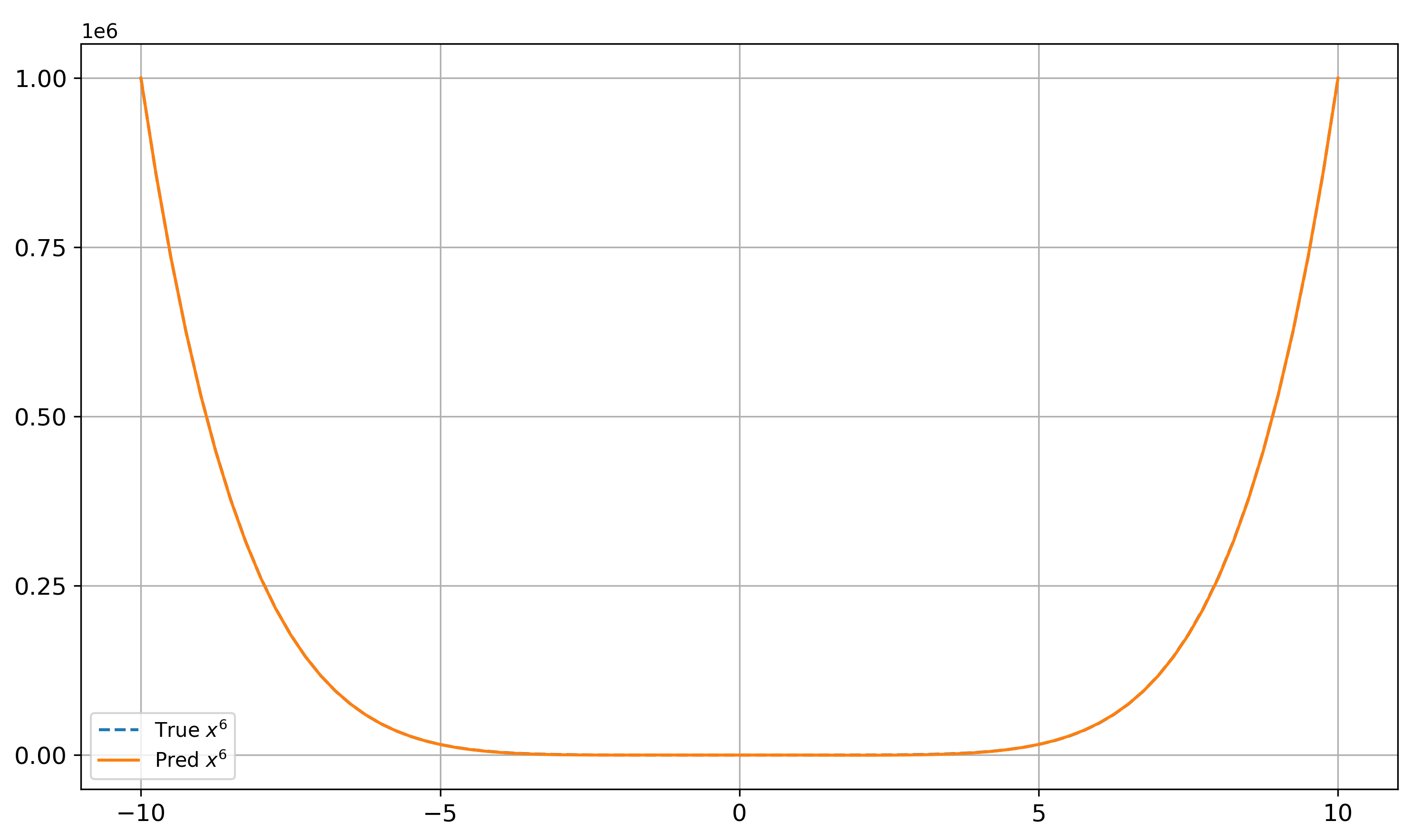}
         \caption*{$f(x) = x^6$}
     \end{minipage}
     \hfill
     \begin{minipage}[b]{0.3\textwidth}
         \includegraphics[width=\textwidth]{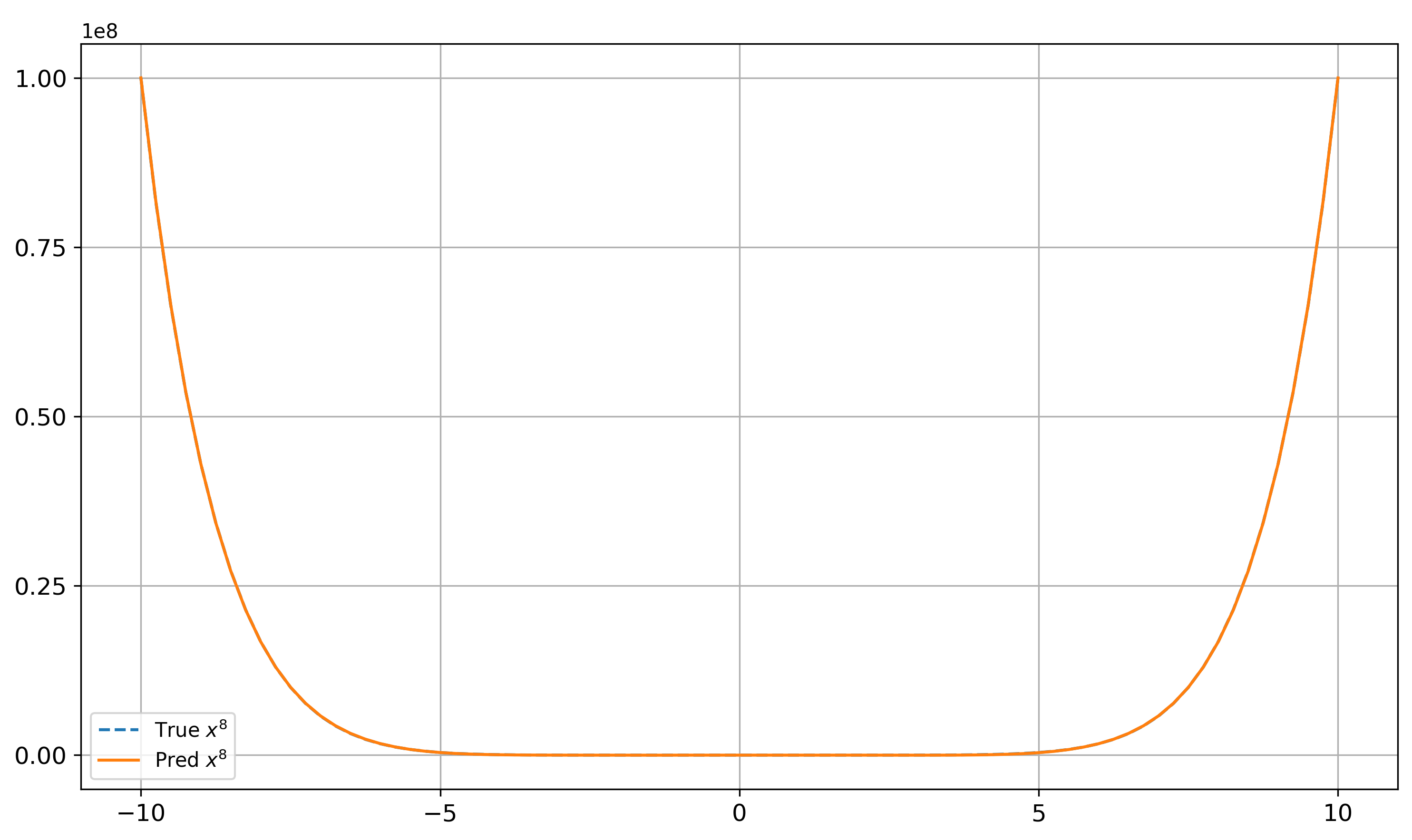}
         \caption*{$ f(x) = x^8 $}
     \end{minipage}

     \caption{Optimal neural network representation of one-dimensional polynomial basis functions.}
     \label{fig:verify_1D}
 \end{figure}

This section investigates the capability of neural networks to represent polynomial basis functions in both one and two dimentions, with a particular focus on out-of-distribution generalization. In one dimention case, we employ a single-hidden-layer neural network to approximate polynomial basis functions of many degrees. The model is trained on data sampled from the interval $[-1, 1]$ and tested on an extended domain $[-10, 10]$. As shown in Figure~\ref{fig:verify_1D}, the predicted values (orange curves) closely match the theoretical polynomials (blue curves), demonstrating the network’s strong extrapolation ability. Notably, the generalization performance is influenced by the complexity of the target function. For higher-degree polynomials, improving model capacity—such as by increasing the number of hidden units or layers—or employing strategies like regularization or piecewise training may enhance extrapolation.

Extending this analysis to two dimention, we consider polynomial basis functions of the form:
$$
f(x_1,x_2) = \sum_{i+j\leq k}c_{ij}x_1^ix_2^j,
$$
where $x_1, x_2 \in \mathbb{R}$, $i,j\geq0$ are integers, and $c_{ij}\in\mathbb R$ are the coefficients. 

\begin{figure}[htb!]
     \centering
     \begin{minipage}[b]{0.3\textwidth}
         \includegraphics[width=\textwidth]{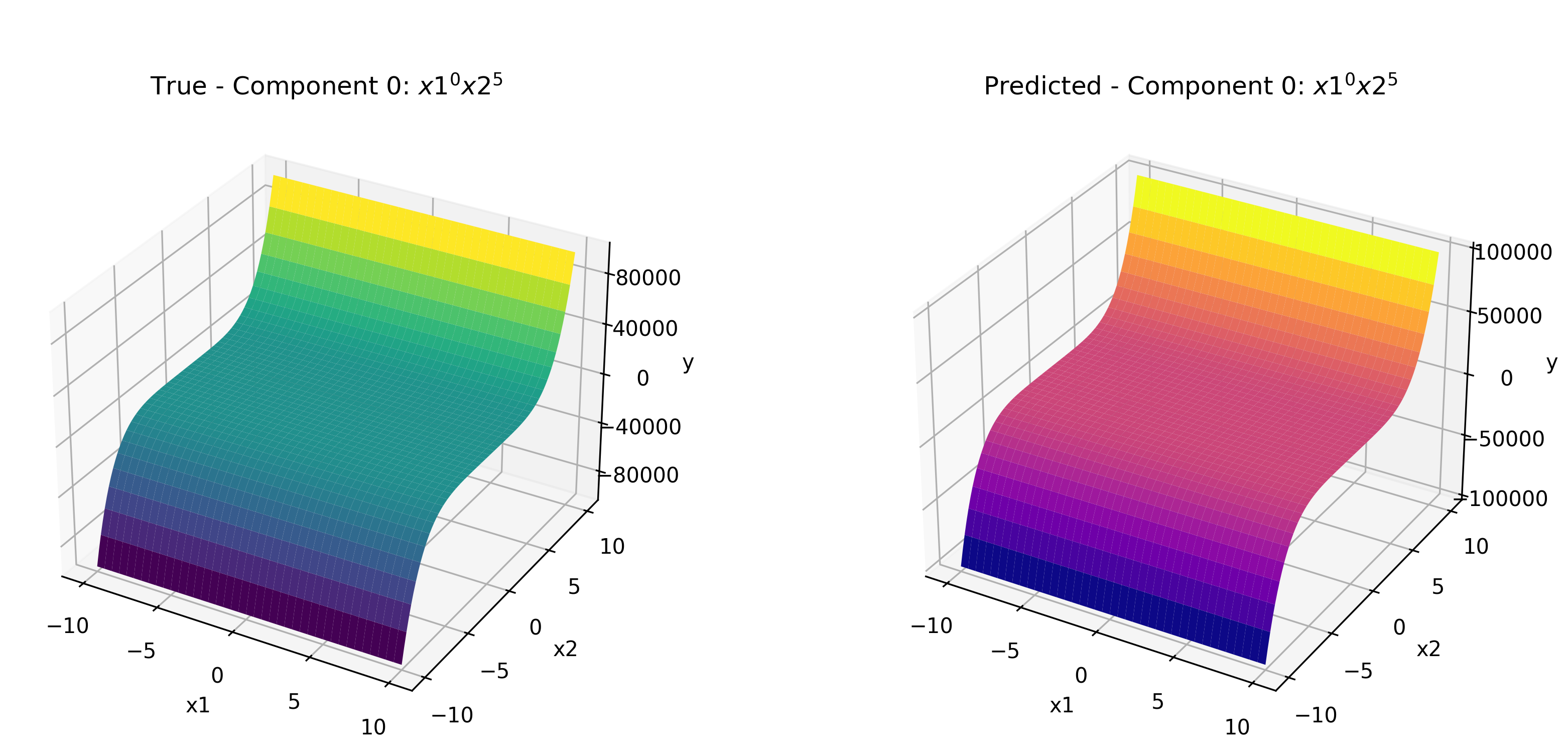}
         \caption*{$f(x) = x_1^0x_2^5$}
     \end{minipage}
     \hfill
     \begin{minipage}[b]{0.3\textwidth}
         \includegraphics[width=\textwidth]{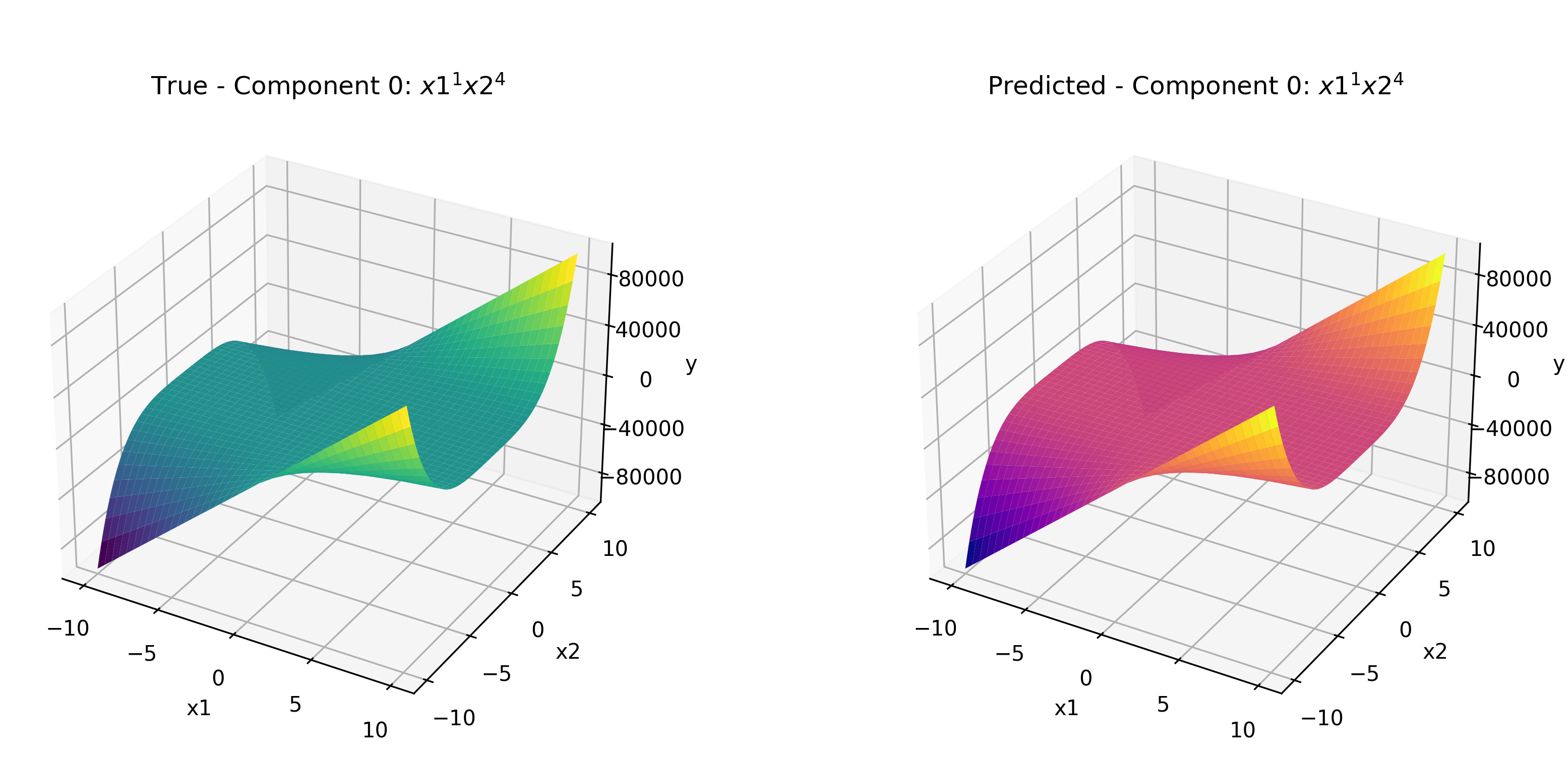}
         \caption*{$f(x) = x_1^1x_2^4$}
     \end{minipage}
     \hfill
     \begin{minipage}[b]{0.3\textwidth}
         \includegraphics[width=\textwidth]{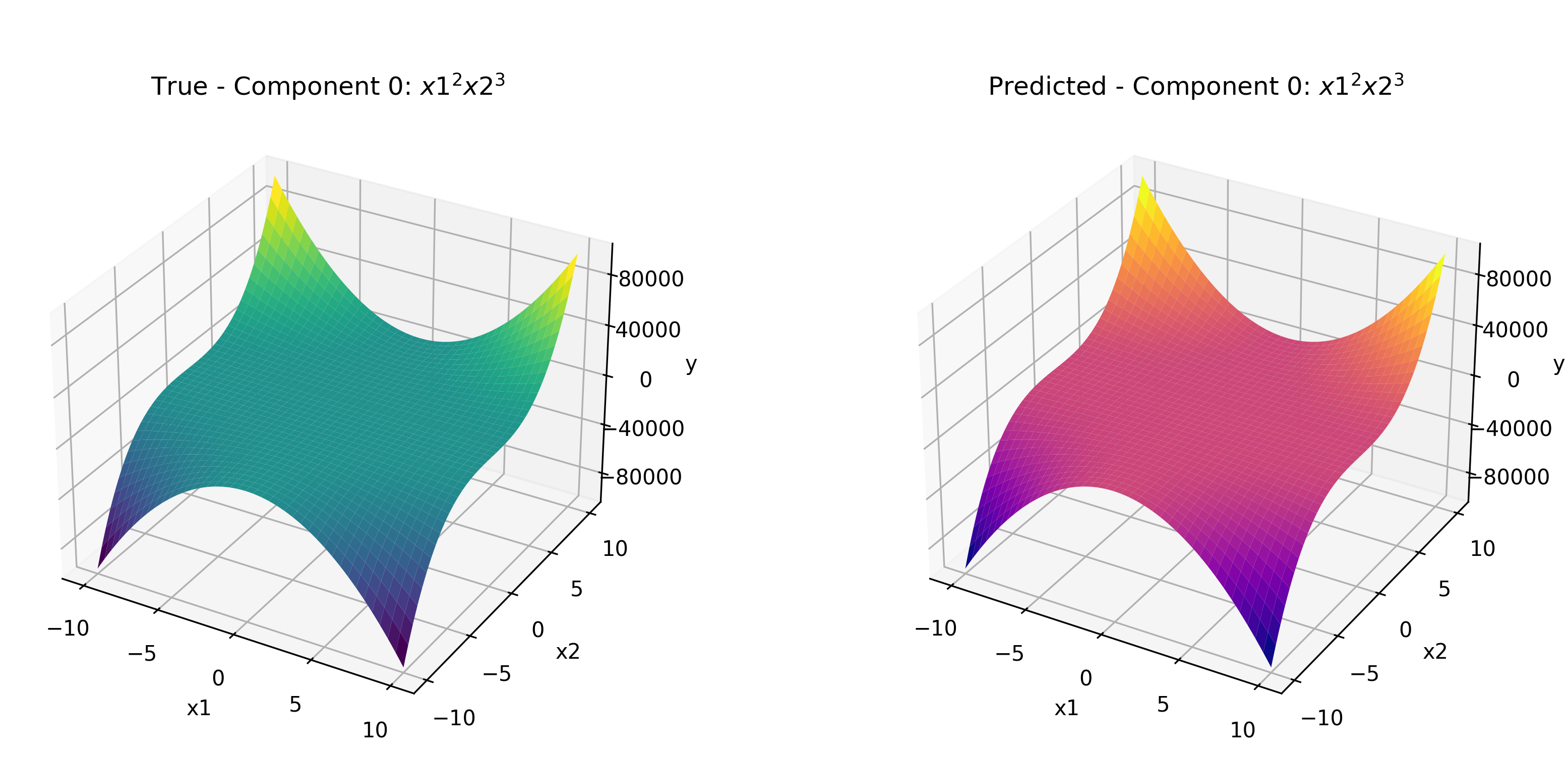}
         \caption*{$f(x) = x_1^2x_2^3$}
     \end{minipage}

     \vspace{0.5cm} 
    
     \begin{minipage}[b]{0.3\textwidth}
         \includegraphics[width=\textwidth]{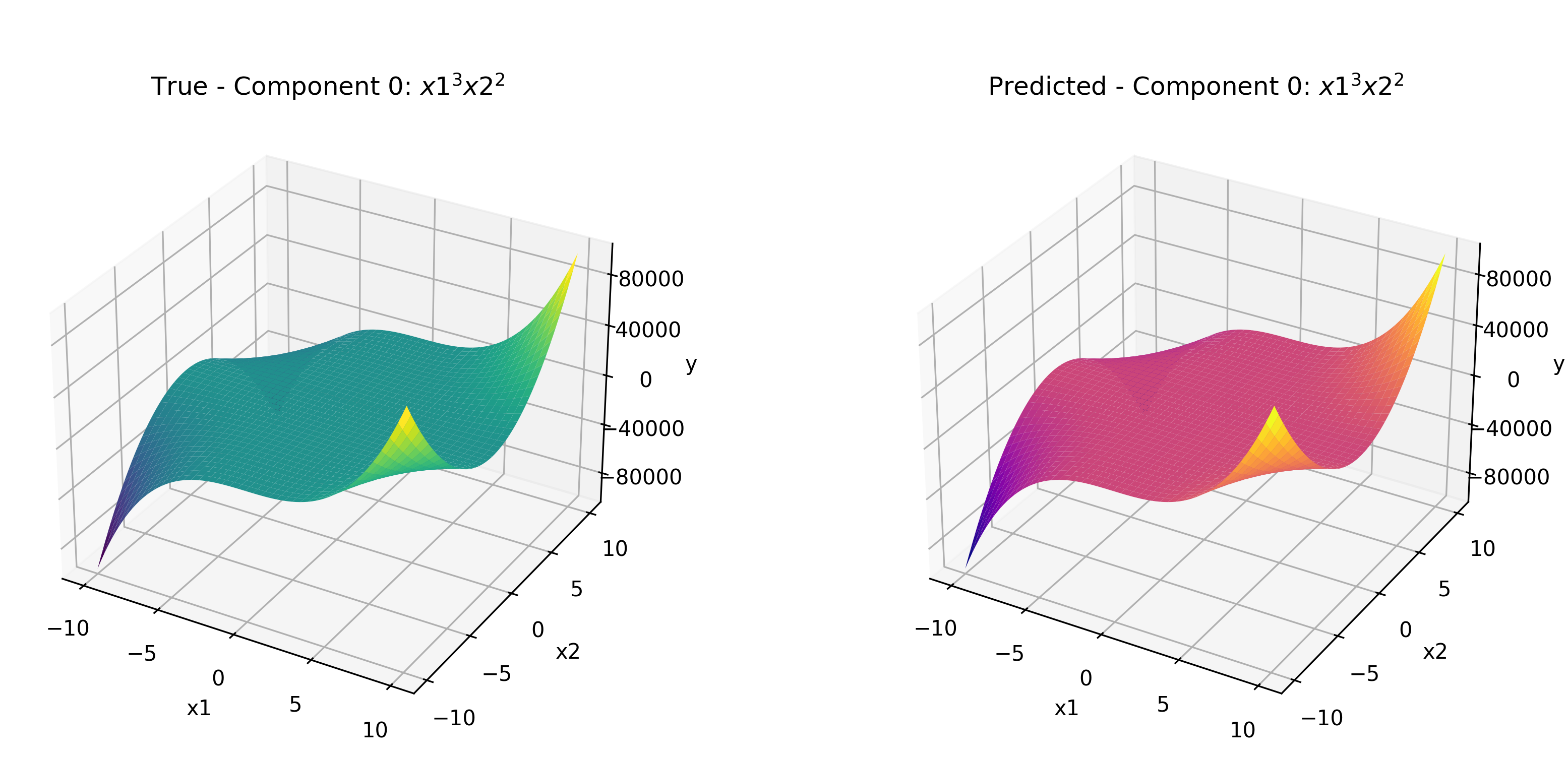}
         \caption*{$f(x) = x_1^3x_2^2$}
     \end{minipage}
     \hfill
     \begin{minipage}[b]{0.3\textwidth}
         \includegraphics[width=\textwidth]{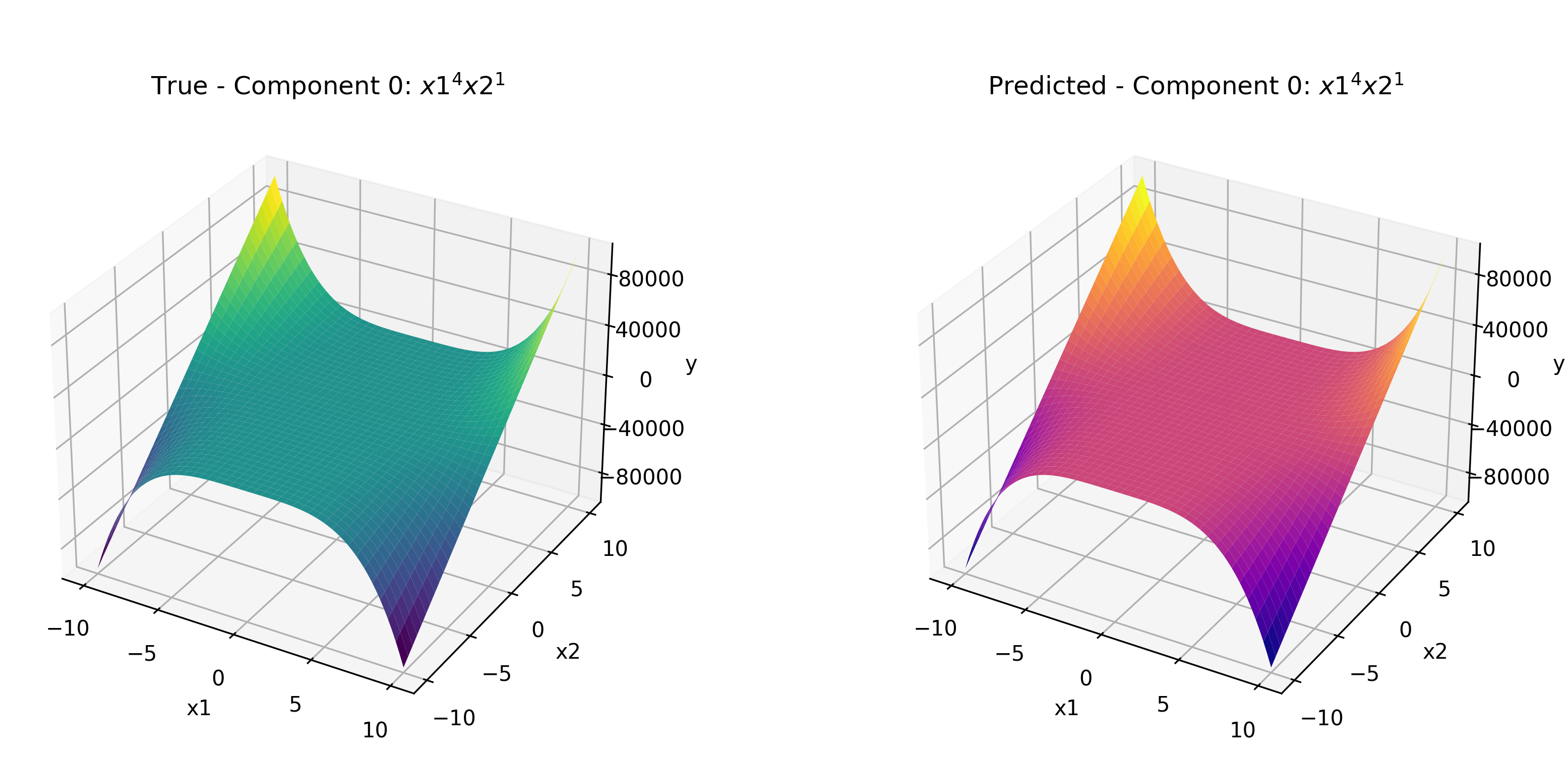}
         \caption*{$f(x) = x_1^4x_2^1$}
     \end{minipage}
     \hfill
     \begin{minipage}[b]{0.3\textwidth}
         \includegraphics[width=\textwidth]{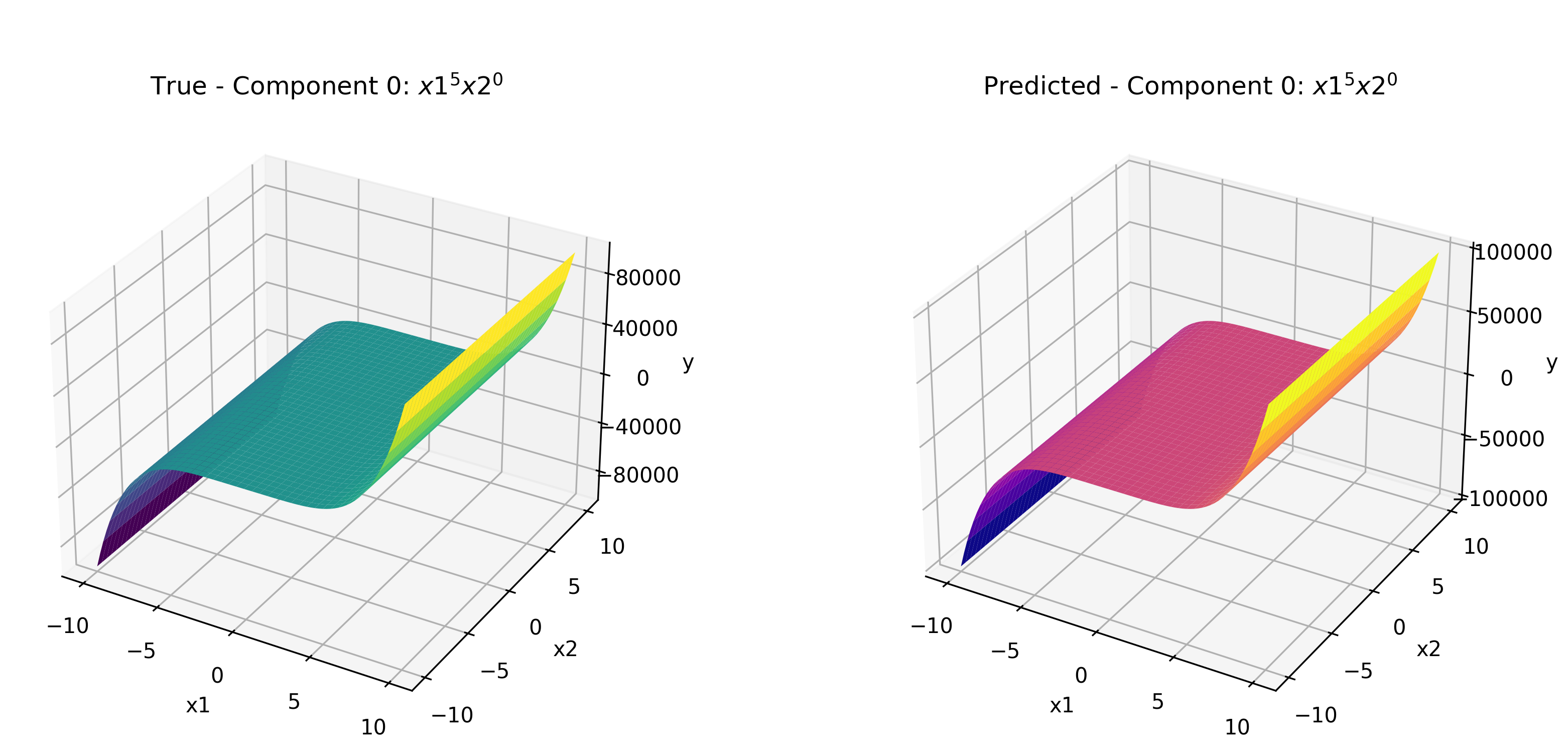}
         \caption*{$f(x) = x_1^5x_2^0$}
     \end{minipage}

    \caption{Optimal neural network representation of two-dimensional polynomial basis functions.}
    \label{fig:verify_2D}
\end{figure}

Each monomial term $x_1^i x_2^j$ is modeled individually using a single neural network. Figure~\ref{fig:verify_2D} displays the predicted outputs for all terms with $i + j = 5$. Similar to one dimention case, the network is trained on $[-1, 1]^2$ and evaluated on the broader domain $[-10, 10]^2$, confirming the network’s ability to generalize polynomial structure to unseen regions. Polynomial approximation examples for various polynomial bases can be found in the publicly available code repository accompanying this paper.

\section{Numerical Experiments}
\subsection{Function Approximation in one dimention}
This section applies the proposed initialization strategy combined with the least squares method to a variety of one-dimensional functions. As shown in Table \ref{tab:metrics_num_exam_1D}, the models consistently achieve extremely low mean squared errors (MSE) and $R^2$ values very close to $1.0$, indicating near-perfect approximation accuracy. These results demonstrate that the proposed approach provides highly reliable approximations across smooth, oscillatory, polynomial, and composite function classes.

\begin{figure}[htb!]
    \centering
    \begin{minipage}[b]{0.3\textwidth}
        \includegraphics[width=\textwidth]{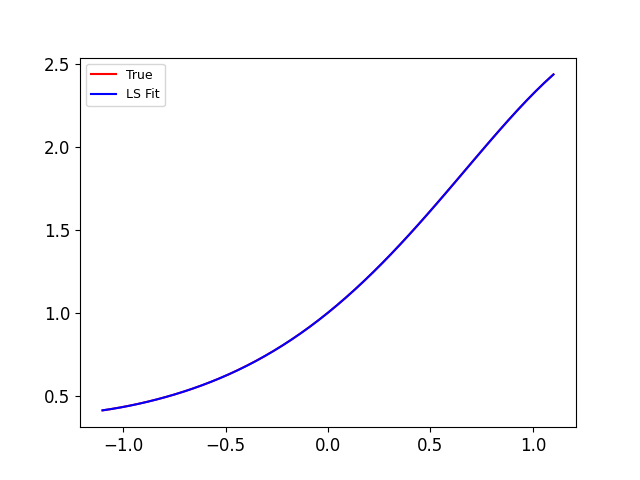}
        \caption*{$f_1(x)$}
    \end{minipage}
    \hfill
    \begin{minipage}[b]{0.3\textwidth}
        \includegraphics[width=\textwidth]{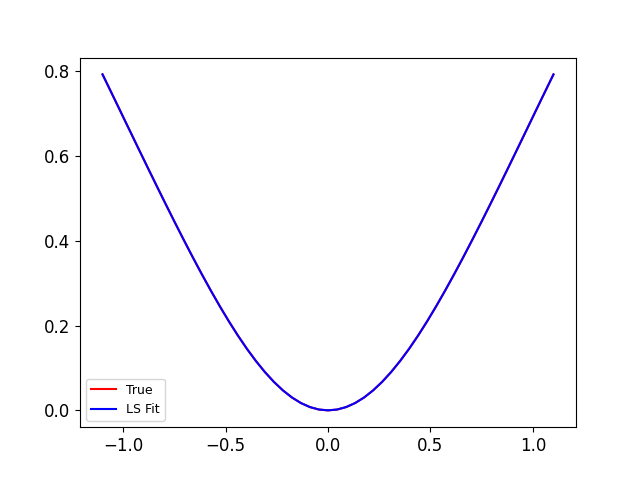}
        \caption*{$f_2(x)$}
    \end{minipage}
    \hfill
    \begin{minipage}[b]{0.3\textwidth}
        \includegraphics[width=\textwidth]{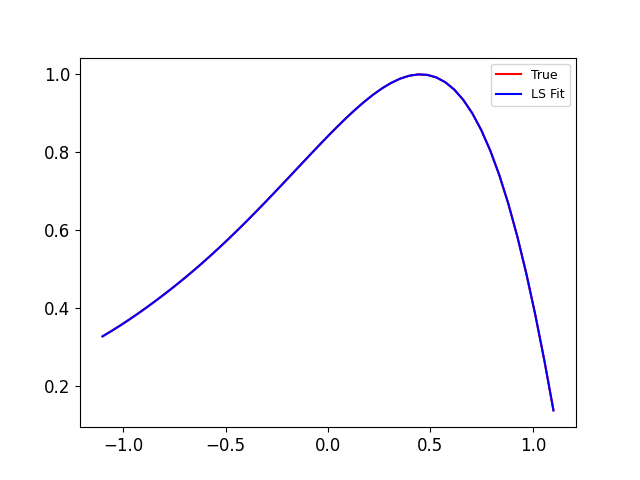}
        \caption*{$f_3(x)$}
    \end{minipage}
    \vspace{0.5cm} 

    \begin{minipage}[b]{0.3\textwidth}
        \includegraphics[width=\textwidth]{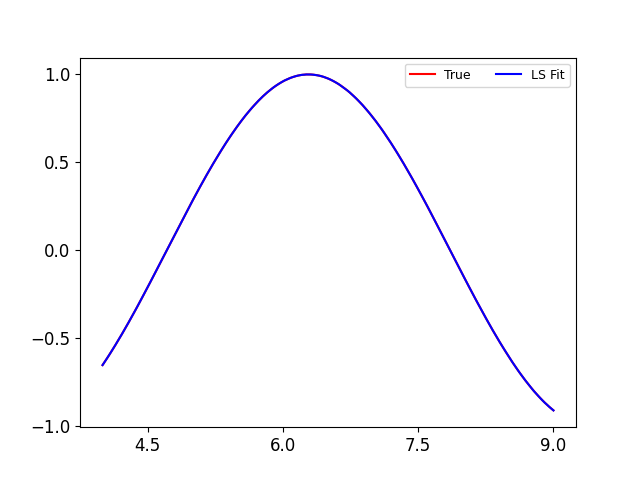}
        \caption*{$f_4(x)$}
    \end{minipage}
    \hfill
    \begin{minipage}[b]{0.3\textwidth}
        \includegraphics[width=\textwidth]{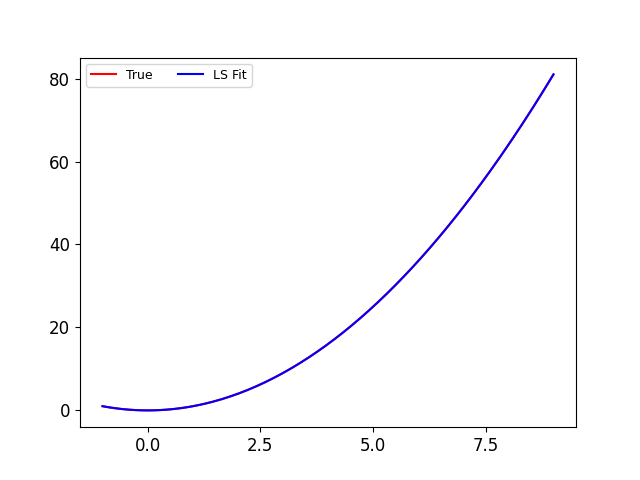}
        \caption*{$f_5(x)$}
    \end{minipage}
    \hfill
    \begin{minipage}[b]{0.3\textwidth}
        \includegraphics[width=\textwidth]{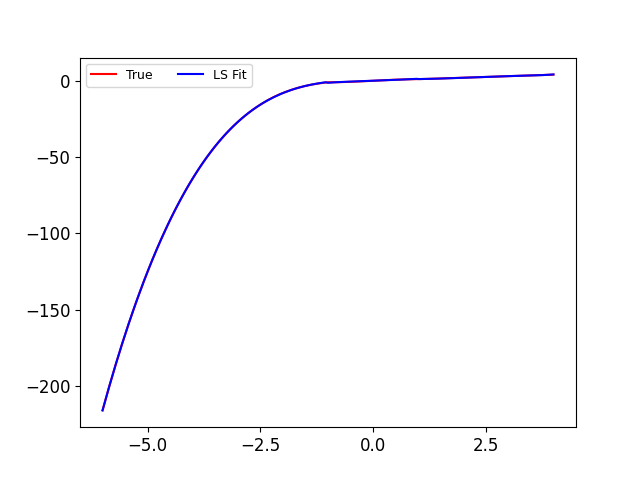}
        \caption*{$f_6(x)$}
    \end{minipage}

    \caption{Visualization of one dimensional function approximation. For each target function listed in Table~\ref{tab:metrics_num_exam_1D}, the fitted curve is compared against the ground truth.}
    \label{fig:groups_approx_1D}
\end{figure}

This conclusion is further supported by the visualization results in Figure \ref{fig:groups_approx_1D}, which illustrates the fitting performance of the six functions in six respective plots. In each plot, the blue line represents the true function, while the red line represents the least squares fitting result. These plots are intended to visually convey the degree of agreement between the true function (True) and the fitted function (LS Fit). In nearly all plots, the red and blue lines completely overlap and are visually indistinguishable, further validating the high precision revealed by the metrics in the table. Even for more complex, piecewise-defined functions, the fitting performance remains strong, demonstrating the robustness of the two-stage fitting approach.

\begin{table}
    \centering
    \caption{Metric results of numerical experiments in one dimension.}\label{tab:metrics_num_exam_1D}
    \begin{tabular}{|l|l|l|l|l|}
    \hline
    functions & max degree & domain & MSE & $R^2$\\\hline
    $f_1(x) = \exp(\sin(x))$ & 6 & $[-1,1]$ & $4.17\times 10^{-8}$ &  $9.9999988\times 10^{-1}$ \\\hline
    $f_2(x) = \ln(1 + x^2)$ & 8 & $[-1,1]$ & $1.05\times 10^{-8}$ &    $9.9999982\times 10^{-1}$\\\hline
    $f_3(x) = \sin(\exp(x))$ & 8 & $[-1,1]$ & $1.64\times 10^{-8}$ &  $9.9999970\times 10^{-1}$\\\hline
    $f_4(x) = \cos(x)$ & 8 & $[4,9]$ & $3.65\times 10^{-8}$ &  $9.9999988\times 10^{-1}$\\\hline
    $f_5(x) = x^2$ & 4 & $[-1,9]$ & $8.39\times 10^{-6}$ &  $1.0$\\\hline
    $f_6(x)=  
        \begin{cases}
        x^3 & \text{if } x < -1, \\
        \sin\left(\frac{\pi x}{2}\right) & \text{if } -1 \leq x < 1, \\
        x & \text{otherwise.}
        \end{cases} $ & 12 & $[-6,4]$ & $5.38\times 10^{-3}$ &  $9.9999827\times 10^{-1}$\\\hline
    \end{tabular}
\end{table} 

\subsection{Function Approximation in two dimentions}
\begin{figure}[htb!]
    \centering
    \begin{minipage}[b]{0.42\textwidth}
        \includegraphics[width=\textwidth]{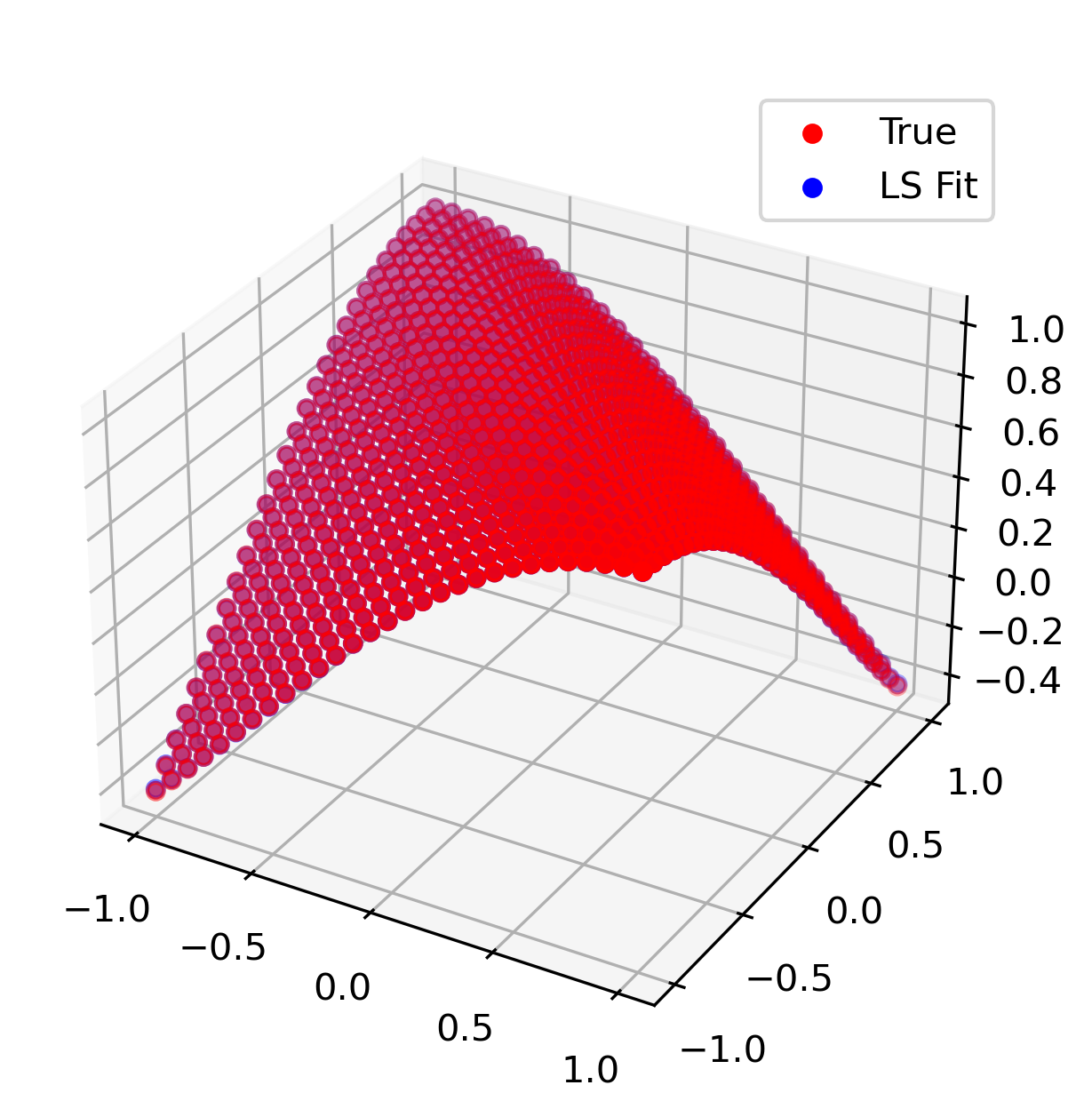}
        \caption*{$f_1(x_1,x_2) = \cos(x_1+x_2)$}
    \end{minipage}
    \hfill
    \begin{minipage}[b]{0.4\textwidth}
        \includegraphics[width=\textwidth]{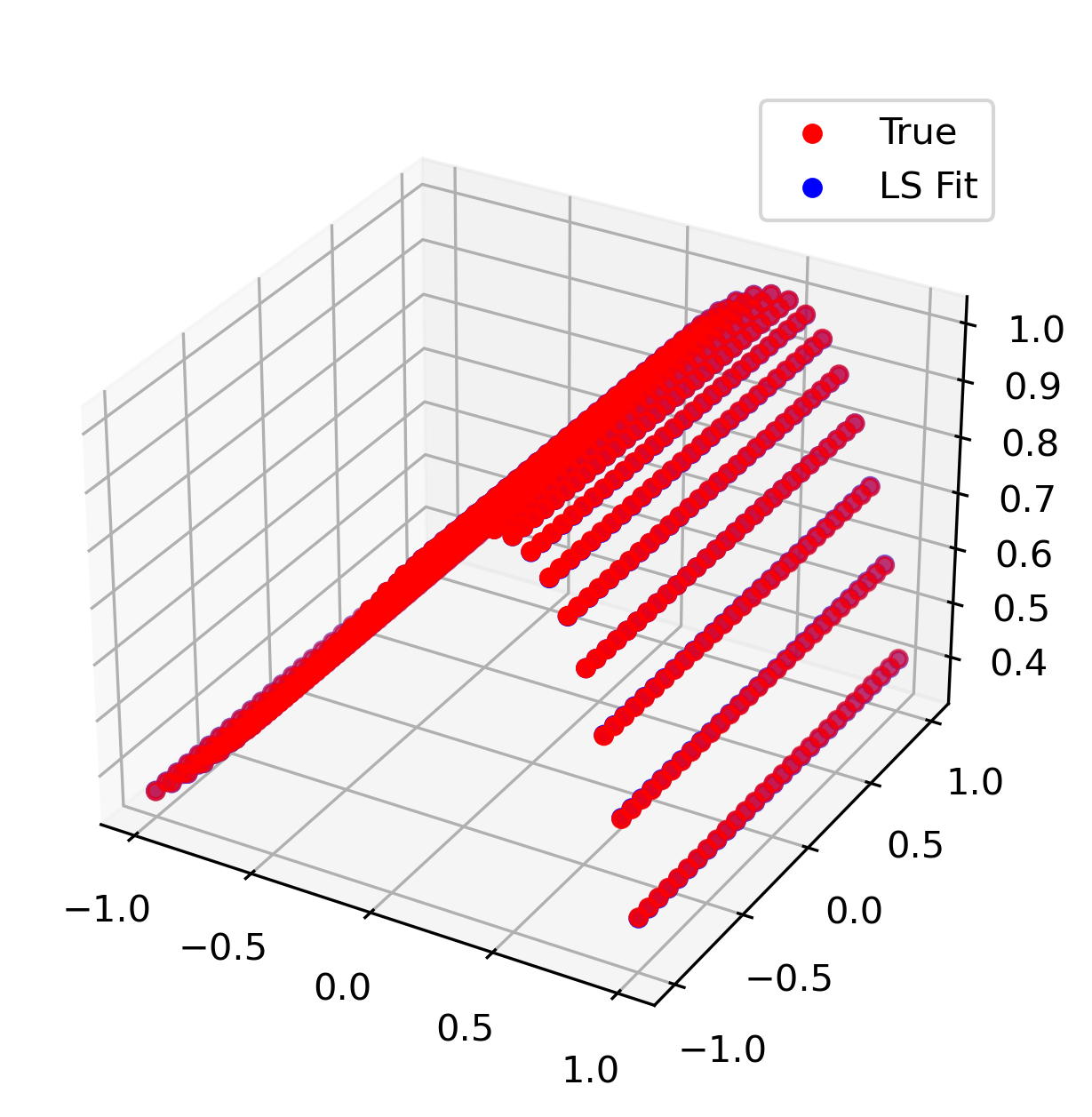}
        \caption*{$f_2(x_1,x_2) = \sin(\exp(x_1))$}
    \end{minipage}

    \vspace{0.5cm} 

    \begin{minipage}[b]{0.4\textwidth}
        \includegraphics[width=\textwidth]{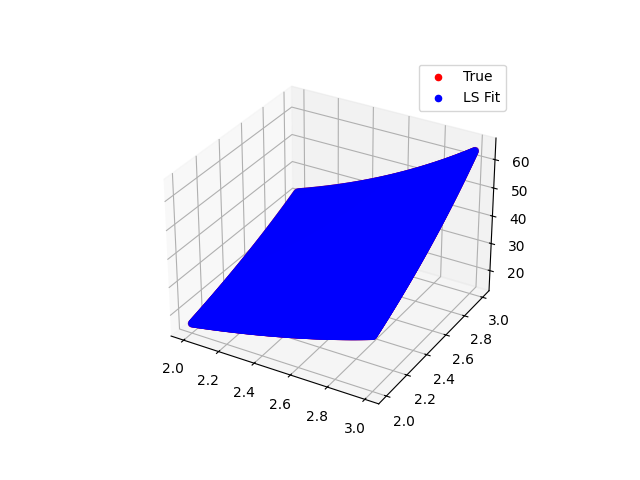}
        \caption*{$f_3(x_1,x_2)=2^{(x_1+x_2)}$}
    \end{minipage}
    \hfill
    \begin{minipage}[b]{0.4\textwidth}
        \includegraphics[width=\textwidth]{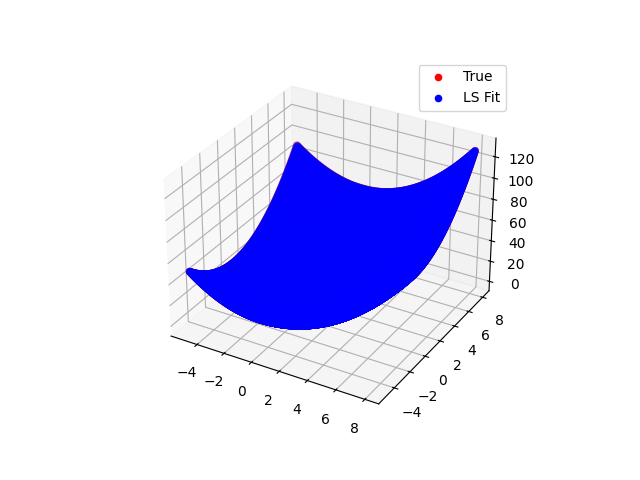}
        \caption*{$f_4(x_1,x_2)=x_1^2+x_2^2$}
    \end{minipage}
    \caption{Visualization of two-dimensional function approximation. For each target function listed in Table~\ref{tab:metrics_num_exam_2D}, the fitted curve is compared against the ground truth.}
    \label{fig:groups_approx_2D}
\end{figure}

Here, we also employ the novel model initialization strategy introduced in this work in conjunction with the least squares approach to approximate general functions over a two dimensional domain, accompanied by quantitative performance metrics and graphical illustrations. The results presented in Figure \ref{fig:groups_approx_2D} and Table \ref{tab:metrics_num_exam_2D}.

For relatively smooth and low-order functions such as the quadratic function
$f_4(x_1, x_2)$, the neural network achieves nearly perfect reconstruction of the target surface. The predicted outputs closely match the ground truth, which is also reflected in the quantitative metrics: the mean squared error (MSE) reaches as low as $8.60 \times 10^{-7}$, with the $R^2$ score attaining $1.0$, indicating excellent predictive accuracy.

For trigonometric-type nonlinear functions, such as
$f_1(x_1, x_2)$ and
$f_2(x_1, x_2)$, despite their oscillatory and composite structures, the networks deliver highly accurate results within the standard domain $[-1,1]^2$. Specifically, the MSE metrics are $2.42 \times 10^{-6}$ and $2.48 \times 10^{-7}$, respectively, with $R^2$ scores of $9.9998331 \times 10^{-1}$ and $9.9995613 \times 10^{-1}$, demonstrating that the models capture both periodicity and nonlinearity with remarkable precision.

Particularly notable is the performance on the exponential-type function
$f_3(x_1, x_2)$, which is defined over a broader domain $[2,3]^2$ and exhibits rapid growth. Even under these conditions, the network maintains strong predictive capability, achieving an MSE of $1.32 \times 10^{-6}$ and an $R^2$ score of $1.0$. This suggests that the model generalizes effectively even for functions with steep gradients and large output ranges.

\begin{table}
    \centering
    \caption{Metric results of numerical experiments in two dimension.}\label{tab:metrics_num_exam_2D}
    \begin{tabular}{|l|l|l|l|l|}
    \hline
    functions & max degree & domain & MSE & $R^2$\\\hline
    $f_1(x) = \cos(x_1+x_2)$ & 4 & $[-1,1]^2$ & $2.42\times 10^{-6}$ &  $9.9998331\times 10^{-1}$ \\\hline
    $f_2(x) = \sin(\exp(x_1))$ & 6 & $[-1,1]^2$ & $2.48\times 10^{-7}$ &    $9.9995613\times 10^{-1}$\\\hline
    $f_3(x) = 2^{(x_1+x_2)}$ & 6 & $[2,3]^2$ & $1.32\times 10^{-6}$ &  $1.0$\\\hline
    $f_4(x) = x_1^2+x_2^2$ & 2 & $[-5,8]^2$ & $8.60\times 10^{-7}$ &  $1.0$\\\hline
    \end{tabular}
\end{table} 

Overall, these results confirm that the proposed weights initialization strategy significantly enhances model performance within standard training regions. Moreover, when combined with the data mapping module, it contributes to improved out-of-domain generalization, highlighting the robustness and adaptability of the method in approximating a variety of function types in two dimensional space.

\section{Conclusion and Future Directions} 
In this work, we presented a neural approximation framework that leverages reusable weights initialization based on pre-trained basis neural networks. By decoupling the representation of polynomial from the learning of target functions, our method facilitates more stable and efficient training while improving approximation accuracy and generalization. Experimental evaluations across a variety of function classes—including polynomials, trigonometric, exponential, and piecewise functions—demonstrated that the proposed initialization strategy consistently outperforms standard randomly-initialized neural networks in both convergence behavior and extrapolation performance.

Several promising directions emerge from this work. One important extension is to higher-dimensional domains, where the proposed initialization strategy may help mitigate the curse of dimensionality through structured basis design or decomposition techniques. Another fruitful direction involves the integration of symbolic priors or physics-informed constraints into the network design and training objectives, enhancing both interpretability and generalization in scientific computing contexts. In addition, the proposed method holds potential for application in operator learning, where the ability to efficiently approximate families of functions or parametric solutions can be significantly improved via reusable basis initialization. Beyond these, a particularly compelling avenue is the extension of this framework to the numerical solution of partial differential equations (PDEs), where the neural approximation of solution functions—guided by the basis-initialized architecture—may offer a robust and flexible alternative to classical discretization methods.

To support this framework, we have developed a modular software package that facilitates key components of the workflow, including data preparation, neural network training, and function approximation. The software is designed with extensibility in mind and includes comprehensive documentation along with an interactive interface for visualization and analysis of approximation results. 
The full source code is openly available at: https://gitee.com/AIxinwen/nn4poly.

\section*{Acknowledgments}

Y. Huang was supported in part by NSFC Project (12431014), Project of Scientific Research Fund of Hunan Provincial Science and Technology Department (2020ZYT003).
N. Yi was supported by the National Key R \& D Program of China (2024YFA1012600) and NSFC Project (12431014).
P. Yin’s research was supported by the University of Texas at El Paso Startup Award.

\section*{Data availability} 
Enquiries about data availability should be directed to the authors.

\section*{Declarations} 
The authors declare that they have no conflict of interest.


\end{document}